\documentclass[10pt]{article}
\usepackage{times}
\usepackage{authblk}
\usepackage{amsmath,amssymb}
\usepackage{graphicx}
\usepackage{hyperref}
\usepackage[margin=1in]{geometry}
\usepackage{microtype}

\usepackage[numbers]{natbib}  % or omit [numbers] if you want author-year style
\usepackage{gensymb}
\usepackage{booktabs}

\title{Uncertainty-Aware Hourly Air Temperature Mapping at 2 km Resolution via Physics-Guided Deep Learning}

\author[1]{Shengjie Kris Liu\thanks{Preprint. Contact emails: \href{mailto:skrisliu@gmail.com}{skrisliu@gmail.com} or \href{mailto:liusheng@usc.edu}{liusheng@usc.edu} (S.K.L.). \href{mailto:siqinwan@usc.edu}{siqinwan@usc.edu} (S.W.)}}
\author[1]{Siqin Wang}
\author[2]{Lu Zhang\thanks{Corresponding author: \href{mailto:lzhang63@usc.edu}{lzhang63@usc.edu} (L.Z.)}}

\affil[1]{Spatial Sciences Institute, Dornsife College of Letters, Arts and Sciences,\protect\\ University of Southern California, Los Angeles, CA, USA}
\affil[2]{Division of Biostatistics, Department of Population and Public Health Sciences, Keck School of Medicine,\protect\\ University of Southern California, Los Angeles, CA, USA}

\date{\today}

\begin{document}
\maketitle

\begin{abstract}
Near-surface air temperature is a key physical property of the Earth's surface. Although weather stations offer continuous monitoring and satellites provide broad spatial coverage, no single data source offers seamless data in a spatiotemporal fashion. Here, we propose a data-driven, physics-guided deep learning approach to generate hourly air temperature data at 2 km resolution over the contiguous United States. The approach, called Amplifier Air-Transformer, first reconstructs GOES-16 surface temperature data obscured by clouds. It does so through a neural network encoded with the annual temperature cycle, incorporating a linear term to amplify ERA5 temperature values at finer scales and convolutional layers to capture spatiotemporal variations. Then, another neural network transforms the reconstructed surface temperature into air temperature by leveraging its latent relationship with key Earth surface properties. The approach is further enhanced with predictive uncertainty estimation through deep ensemble learning to improve reliability. The proposed approach is built and tested on 77.7 billion surface temperature pixels and 155 million air temperature records from weather stations across the contiguous United States (2018–2024), achieving hourly air temperature mapping accuracy of 1.93\degree C in station-based validation. The proposed approach streamlines surface temperature reconstruction and air temperature prediction, and it can be extended to other satellite sources for seamless air temperature monitoring at high spatiotemporal resolution. The generated data of this study can be downloaded at \url{https://doi.org/10.5281/zenodo.15252812} and the project webpage can be found at \url{https://skrisliu.com/HourlyAirTemp2kmUSA/}. 
\end{abstract}

\section{Introduction}
\label{sec:intro}
Near-surface air temperature refers to the temperature of air measured at a height of 2 m above the Earth's surface~\citep{prihodko1997estimation}. It is the temperature record most closely related to living environments, playing a critical role in cities, rural areas, ecosystems and beyond~\citep{lobell2007changes, liu2020local, lee2020projections}. As the near-surface air temperature, it is sensitive to Earth's surface materials and land cover, varying across locations~\citep{winckler2018different, cao2021within} and can have large diurnal variation within one day~\citep{makowski2008diurnal, liu2023spatial, liu2024effects}. It is therefore critical to have near-surface air temperature data at high spatiotemporal resolution. 

Historically collected through meteorological stations, near-surface air temperature data are known to be sparsely distributed in space due to the high set up and maintenance cost~\citep{fan2008global}. Most meteorological stations are maintained within airports, often one or two within a city~\citep{krishnan2015comparison}. Due to this limitation, research on complex societal issues often simplifies their hypothesis and assumes the same temperature value within a city~\citep{liu2024effects, barnett2010measure, gasparrini2015mortality}. Satellites, as the other often-used data source, offer broad spatial coverage at low cost, but this broad spatial coverage is still limited by weather conditions, especially clouds~\citep{li2013satellite}. Even without clouds, most satellites can only provide one to two data snapshots per day because of their polar orbits~\citep{wu2015integrated, yoo2018estimation, jia2024advances}. 

One type of satellite, the geostationary satellite, can provide temperature data throughout the day. With improved sensor technology, the GOES-16 satellite series, the latest in geostationary satellites, has enabled temperature monitoring at 2 km resolution, with observations ranging from hourly to every five minutes since 2016~\citep{beale2019comparison}. However, only a limited number of studies have utilized this data source to investigate hourly near-surface air temperature dynamics~\citep{hrisko2020urban, zhang2022hourly}, with only a few focusing on hourly land surface temperature completion and analysis~\citep{jia2022generating, pestana2022evaluating, liu2025resolution}. Although earlier work has demonstrated the potential of using GOES-16 data for hourly air temperature mapping, cloud cover frequently causes missing values, hindering complete estimation of near-surface air temperature~\citep{zhang2022hourly}. Generating near-surface air temperature data requires addressing two challenges simultaneously: filling in surface temperature gaps caused by cloud cover and estimating air temperature from surface temperature. These two processes are typically handled separately in existing approaches.

Clouds are the major obstacle to use satellite temperature data. Many methods have been developed for land surface temperature (LST) completion (or called reconstruction), primarily focusing on using statistics or energy balanced models~\citep{jia2024advances, wu2021spatially}. Many spatiotemporal reconstruction methods have been proposed, including spatiotemporal gap filling and energy-balanced calculation~\citep{liu2017spatiotemporal}. While LST data are fundamental and often are the foundation to generate other types of temperature data, near-surface air temperature data are critically needed in real-world applications~\citep{white2013validating}. However, the transformation from surface temperature to air temperature is not trivial, and the validation on LST data has been limited by the even lesser number of in-situ LST sites, e.g., only six from the Surface Radiation Budget Network within the United States~\citep{augustine2000surfrad}. Another problem is that in-situ radiation sites can only be set up in remote areas with natural land cover, making validation impossible in complex urban areas~\citep{jia2024advances}. The lack of in-situ LST sites and the critical need of air temperature data have driven a new validation scheme for LST reconstruction in recent years, which uses the capability to generate near-surface air temperature as an indirect validation for LST reconstruction~\citep{zhao2020reconstruction, yang2024annual, 2025arXiv250214433L}. Still, since transforming LST to air temperature is non-trivial, existing approaches limit comparisons to the same station only, separately for original and reconstructed LST. It remains challenging to estimate air temperature on a large scale.

Estimating near-surface air temperature from surface temperature adds another layer of complexity. Satellite observations capture blackbody radiance---an energy quantity—--which can be converted to surface temperature with additional information such as emissivity and atmospheric conditions~\citep{becker1995surface}. Once the Earth's surface is heated, the temperature difference between the surface and the near-surface air drives heat transfer, warming the lower atmosphere through conduction~\citep{emanuel1994atmospheric}. As the heated air becomes less dense, it rises, transferring heat upward via convection. These processes are influenced by various factors, including atmospheric moisture content and air density~\citep{emanuel1994atmospheric}. Due to the presence of a temperature gradient, surface temperature and near-surface air temperature are related but distinct, sparking critical discussions on the appropriate usage of various forms of temperature data~\citep{li2013satellite, jia2024advances}. Nonetheless, as surface temperature is the driving factor, it is possible to estimate near-surface air temperature from surface temperature given sufficient information. To this end, machine learning models are key to transforming surface temperature to near-surface air temperature. Existing frameworks include using random forest for hyperlocal air temperature mapping within a city in Norway~\citep{venter2020hyperlocal}. In the study with GOES-16 data, random forest is also used to estimate near-surface air temperature data from surface temperature~\citep{zhang2022hourly}. These methods often result in model accuracy at around 2-3\degree C for daily estimates~\citep{shen2020deep}. Monthly average estimate can reach to 1.5\degree C~\citep{hooker2018global}. 

\begin{figure}[htbp]
    \centering
    \includegraphics[width=0.6\linewidth]{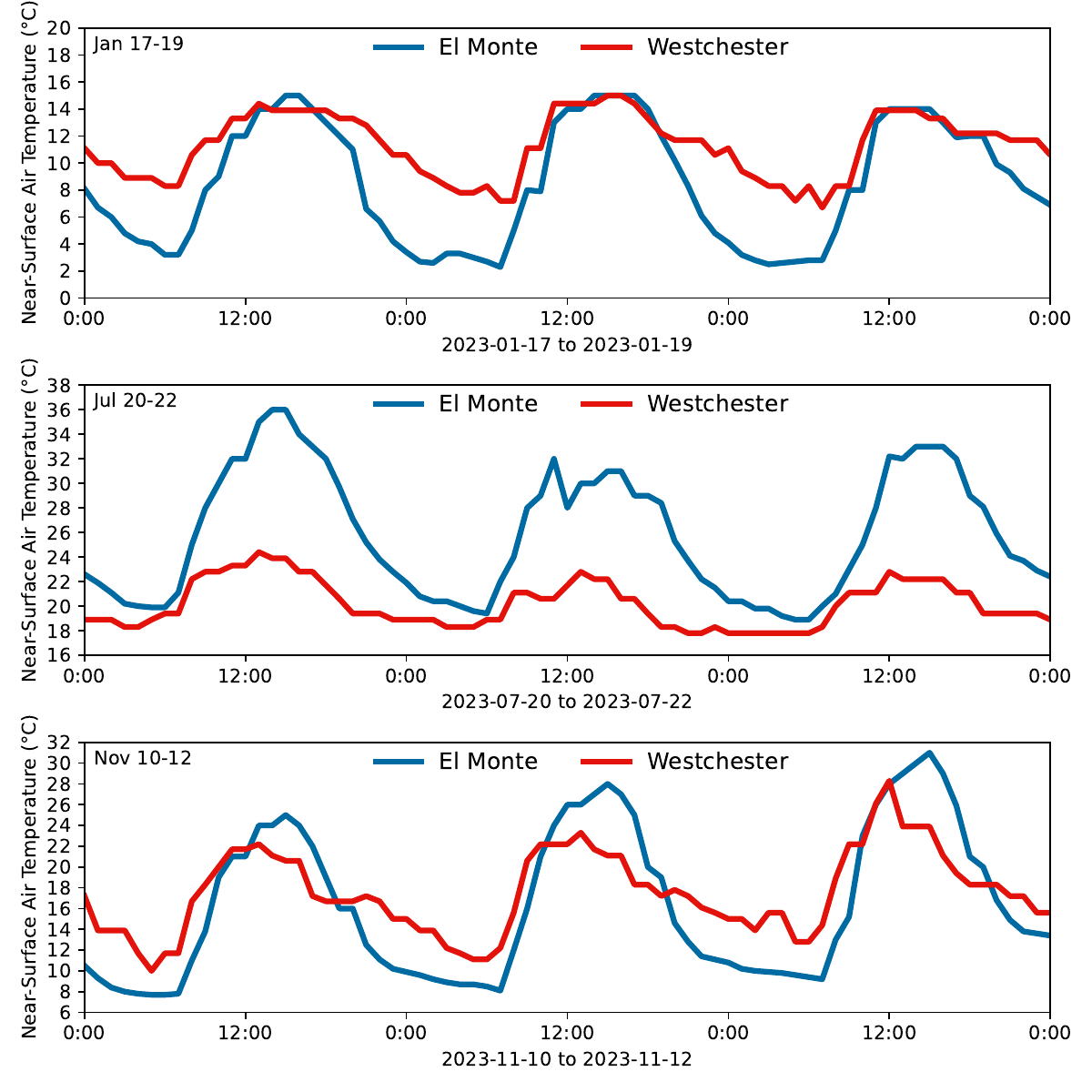}
    \caption{Large spatiotemporal air temperature variations between two meteorological stations, El Monte and Westchester, located approximately 30 km apart within the Los Angeles metropolitan area.}
    \label{fig:SampleLA}
\end{figure}

Estimating sub-daily air temperature is crucial but remains underexplored in existing literature, primarily due to the limited availability of high-frequency temperature observations. Air temperature can fluctuate significantly within a single day, even across short distances. For instance, two meteorological stations just 30 km apart within the Los Angeles metropolitan area can exhibit substantial variations, highlighting the dynamic nature of air temperature changes across both space and time (Figure~\ref{fig:SampleLA}). Intra-day surface and air temperature fluctuations play a vital role in human society and ecosystems~\citep{lambrechts2011impact, meier2019biomass, lee2018mortality}, and they also serve as a key indicator of global climate trends~\citep{braganza2004diurnal}. Capturing hourly air temperature variations at high spatial resolution is therefore essential for deepening our understanding of these complex dynamics.

In this study, we propose a data-driven, physics-guided deep learning approach to generate gap-free, hourly near-surface air temperature data across the contiguous United States (2018--2024). The proposed approach, consisting of two neural networks, is called the Amplifier Air-Transformer. First, the Amplifier model reconstructs surface temperature obscured by clouds in GOES-16 data using a convolutional neural network (CNN) encoded with the annual temperature cycle. It incorporates an additional linear term to amplify ERA5 temperature values (0.1\degree\ resolution) to the resolution of GOES-16 (2~km), with convolutional layers capturing the remaining spatiotemporal variations~\citep{liu2025resolution}. Second, another neural network transforms the reconstructed surface temperature into near-surface air temperature by leveraging its latent relationship with key Earth surface and near-surface properties. The use of two neural network models greatly enhances the exploration of patterns in millions of air temperature records and billions of surface temperature pixels. The proposed approach is further enhanced through deep ensemble learning, which provides prediction intervals and quantifies uncertainty in the generated near-surface air temperature estimates.

In the remainder of this paper, we first introduce the study area and the data, including GOES-16 LST and near-surface air temperature from meteorological stations, in Section~\ref{sec:StudyAreaData}. We then detail the proposed approach in Section~\ref{sec:method}, followed by results and analysis in Section~\ref{sec:ResultAnalysis}. Finally, we present the discussion and conclusion in Section~\ref{sec:Conclusion}.

\section{Study area and data}
\label{sec:StudyAreaData}
\subsection{Study area}

\begin{figure*}[htbp]
    \centering
    \includegraphics[width=0.9\linewidth]{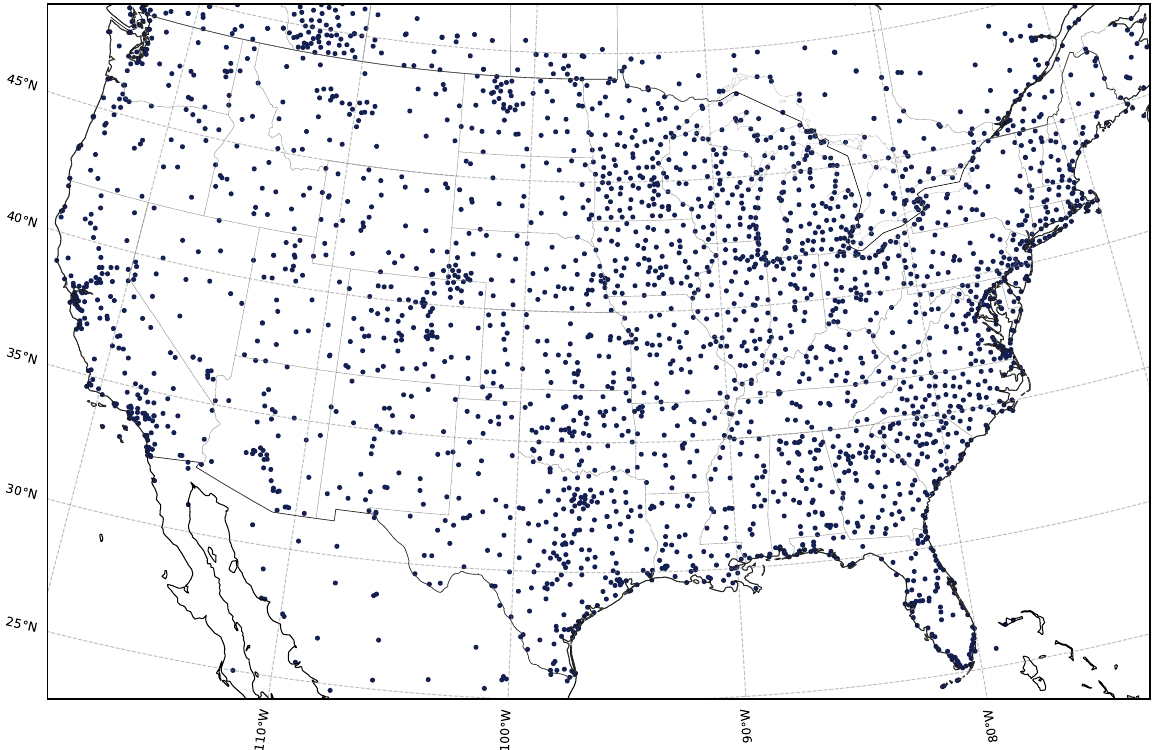}
    \caption{Study area, and the meteorological stations across the contiguous United States.}
    \label{fig:map0_stations}
\end{figure*}

The study area covers the contiguous United States (CONUS), including the lower 48 states and a small portion of adjacent regions in Canada and Mexico for analytical continuity (Figure~\ref{fig:map0_stations}). A no-data mask is applied to pixels lacking valid GOES-16 LST observations, primarily over the ocean. As inland water bodies and land-water boundary areas contain valid surface temperature data, these areas are retained. 

\subsection{In situ near-surface air temperature}
In situ near-surface air temperature data (2018–-2024) were obtained from the Integrated Surface Dataset (ISD) via NOAA~\citep{smith2011integrated}. This global, hourly dataset provides near-surface air temperature observations from meteorological stations worldwide. Stations within the study area were first selected, and then those without 50\% valid observations each year were discarded for the specific year. Due to the addition and discontinuation of some stations, the number of valid meteorological stations varies annually, ranging from 2,614 to 2,658. In total, 155,144,750 valid air temperature records are available in 2018-2024. The number of stations and corresponding valid air temperature records are summarized in Table~\ref{tab:samples}. An overall increasing trend in data availability is observed, except for the latest year (2024), likely due to a lag in data entries as the dataset used in this study was downloaded in early 2025.

\begin{table}[htbp]
  \centering
  \caption{Near-surface air temperature data: number of valid stations and records.}
    \begin{tabular}{lll}
    \toprule
    Year  & \# Stations & \# Records of Air Temperature \\
    \midrule
    2018  & 2,636  & 22,248,310 \\
    2019  & 2,614  & 22,065,224 \\
    2020  & 2,623  & 22,179,774 \\
    2021  & 2,638  & 22,144,810 \\
    2022  & 2,652  & 22,292,766 \\
    2023  & 2,658  & 22,320,213 \\
    2024  & 2,638  & 21,893,653 \\
    \bottomrule
    \end{tabular}%
  \label{tab:samples}%
\end{table}%

\subsection{GOES-16 land surface temperature}
Hourly land surface temperature (LST) data were obtained from the GOES-16 geostationary satellite. The LST data exclude areas obstructed by clouds. We collected all available records from 2018–-2024 via NOAA through Amazon Web Services (AWS). Formally known as the GOES-R Advanced Baseline Imager (ABI) Land Surface Temperature product, the dataset is derived using the split-window technique from two ABI thermal bands: 11.2~\textmu m and 12.3~\textmu m~\citep{schmit2018applications}. Over the CONUS region, the data have a 2~km spatial resolution, varying slightly due to differences in field of view. The retrieved LST product has a precision of 2.3~K and an accuracy of 2.5~K under conditions of known surface emissivity and proper atmospheric correction; when these conditions are not met, accuracy decreases to 5~K~\citep{schmit2018applications}. Only cloud-free pixels contain valid LST retrievals~\citep{schmit2018applications}.

\subsection{GOES-16 spectral reflectance}
To characterize surface properties, we utilized surface reflectance data from the GOES-16 ABI products. Five spectral bands were selected: 0.47~\textmu m, 0.64~\textmu m, 0.86~\textmu m, 1.61~\textmu m, and 2.26~\textmu m. These bands span the visible and near-infrared wavelengths, offering comprehensive information on land surface characteristics. Valid daytime observations were used to compute the annual mean spectral reflectance.

\subsection{Surface temperature at coarse resolution}
We obtained coarse-resolution surface temperature data from the European Centre for Medium-Range Weather Forecasts (ECMWF) Reanalysis version 5 (ERA5). Specifically, we used the ERA5-Land hourly skin temperature product, which has a 0.1\degree~spatial resolution. The ERA5 skin temperature data were matched to the nearest GOES-16 LST pixels based on spatial proximity.

\subsection{Input features for transforming surface temperature to air temperature}
\label{ssec:features}
Additional input data are obtained for transforming surface temperature to near-surface air temperature. These data, listed in Table~\ref{tab:features}, are categorized into three types: spatiotemporal indices (latitude, longitude, hour of the day), spatially varying features (DEM, slope), and hourly reanalysis data at coarse resolution (boundary layer height, total column water, instantaneous surface sensible heat flux, 10-meter U-component of wind, and 10-meter V-component of wind).

\begin{table}[htbp]
  \centering
  \caption{Features used to transform surface temperature to air temperature}
    \begin{tabular}{ll}
    \toprule
    \textbf{Category} & \textbf{Features} \\
    \midrule
    Primary      & Surface Temperature \\
                 & Surface Reflectance \\   
    Spatiotemporal Index & Latitude \\
          & Longitude \\
          & Hour of Day \\
    Spatial Varying Feature & Elevation \\
          & Slope \\
    Hourly Reanalysis Data & Boundary Layer Height \\
          & Total Column Water \\
          & Surface Sensible Heat Flux \\
          & Zonal Wind (u10) \\
          & Meridional Wind (v10) \\
    \bottomrule
    \end{tabular}%
  \label{tab:features}%
\end{table}%

Adding spatiotemporal indices, including latitude, longitude, and hour of the day, enables the model to explicitly capture spatiotemporal signals. Elevation is obtained from the NASA NASADEM Digital Elevation 30~m product, which was enhanced from the Shuttle Radar Topography Mission (SRTM) with additional data from the Terra Advanced Spaceborne Thermal and Reflection Radiometer (ASTER) Global Digital Elevation Model (GDEM), ICESat Geoscience Laser Altimeter System, and Advanced Land Observing Satellite Panchromatic Remote-sensing Instrument for Stereo Mapping (PRISM) AW3D30 DEM~\citep{simard2024global}. With an RMSE of 1.5~m, NASADEM represents the current state-of-the-art digital elevation model~\citep{simard2024global}. Slope is then calculated from the DEM data. Aspect is not used, as an initial test indicated that aspect degraded the algorithm's performance.

Five hourly ERA5 reanalysis variables at 0.25\degree\ resolution are included. These variables capture changing atmospheric conditions that influence heat transfer from the surface to the near-surface air. Boundary Layer Height (BLH) represents the thickness of the lowest layer of the atmosphere in direct contact with the Earth's surface, indicating the degree of turbulent mixing and ventilation, which impacts surface heat transfer to the air~\citep{chen2022lidar}. Total Column Water (TCW) quantifies the total amount of water vapor in the atmosphere, affecting surface-air heat exchange by influencing long-wave radiation~\citep{loew2016high}. Surface Sensible Heat Flux (SHF) measures the transfer of heat from the Earth's surface to the atmosphere and is directly related to surface-air temperature transformation~\citep{jia2003estimation}. Finally, zonal wind (u10) and meridional wind (v10) capture the intensity and direction of west-east and north-south winds, respectively, which affect turbulent mixing.

\section{Methodology}
\label{sec:method}
\subsection{Problem statement and method overview}
The goal of this study is to generate hourly near-surface air temperature data from incomplete surface temperature data. This goal includes two processes: reconstructing surface temperature, and transforming surface temperature to air temperature. Here, we propose a data-driven, physics-guided deep learning approach called Amplifier Air-Transformer. 

\subsubsection{Reconstruct surface temperature}
First, the Amplifier model, which is a physics-guided convolutional neural network encoded with the annual temperature cycle, is used to reconstruct GOES-16 surface temperature. An additional linear term built in the neural network is used to amplify the coarse-resolution ERA5 surface temperature (0.1\degree\ resolution) to the GOES-16 data's resolution (2~km), hence the name Amplifier. After the annual temperature cycle and the linear term, several convolutional layers are used to model the remaining spatiotemporal variations. As ERA5 is a climate reanalysis product supported by fluid dynamics and thermodynamics, the Amplifier model in the proposed approach is therefore bounded by this restriction. And, it is also bounded by the GOES-16 observations, which enables it to generate surface temperature nearly identical to the observations. These elements make the model data-driven and guided through physical processes. 

\subsubsection{Transform surface temperature to air temperature}
After obtaining the reconstructed surface temperature, a second neural network model of the approach, is used to transform the reconstructed surface temperature to near-surface air temperature. This is through using neural networks to identify the underlying latent relationship among surface temperature, air temperature, and surface and near-surface properties. 

\subsubsection{A big data problem and modern neural networks as the solution}
Due to hourly estimation and large spatial coverage, there are about 12 billion data pixels and 22 million air temperature data records each year, totaling to 77.7 billion surface temperature pixels and 155 million air temperature records in this study. As such, classic machine learning models, such as random forest with a complexity of \(  \mathcal{O}\left(n \log\left(n\right)\right) \) and support vector machines with a complexity of \( \mathcal{O}\left(n^3\right) \), become insufficient and even impossible to handle generating hourly temperature data~\citep{oshiro2012many, abdiansah2015time}.  Modern neural networks have been proven efficient in capturing complex latent relationships, with a linear complexity of  \( \mathcal{O} \left(n\right) \), and have been extensively accelerated through GPU-accelerated computing~\citep{zhang2016deep, zhu2017deep, liu2020few}. By leveraging the advantages of modern neural networks, the proposed approach can effectively reconstruct surface temperature and facilitates its transformation to air temperature with high accuracy.

\subsection{The Amplifier Air-Transformer approach}
We denote the two neural networks in the Amplifier Air-Transformer approach as \( \mathcal{A} \)  and \( \mathcal{T} \), respectively, and represent the combined model as \( \mathcal{M} = \{ \mathcal{A}, \mathcal{T} \} \). Let \( T_\text{surf} \) be the surface temperature and \( T_\text{air} \) the near-surface air temperature. The goal is to train two neural networks within the proposed approach to estimate air temperature from surface temperature at time \( t \), along with additional variables \( X_t \), which include surface and near-surface properties. The model is defined as:
\begin{equation}
T_\text{air} = \mathcal{M} \left( T_\text{surf}, t, X_t \right) = \mathcal{T} \circ \mathcal{A} \left( T_\text{surf}, t, X_t \right) \;.
\end{equation}

\subsubsection{The Amplifier: physics-guided deep learning}
The Amplifier model is a neural network encoded with annual temperature cycle and a linear term to amplify coarse-resolution ERA5 temperature data~\citep{liu2025resolution}. Given incomplete GOES-16 surface temperature \( T_\text{surf} \), we have the Amplifier model as, 
\begin{equation}
\hat{T}_\text{surf}^{\mathcal{A}} \left(t\right) = \left[ \mathcal{T}_{\text{ATC}}(t) + \rho T_{\text{coarse}}(t) \right]  + \mathcal{N}_{conv} \left(X_t\right) \;, 
\end{equation}
where \( \hat{T}_\text{surf}^{\mathcal{A}} \) is the reconstructed surface temperature, \( \mathcal{T}_{\text{ATC}}\left(t\right) \) the annual temperature cycle component as a function of \( t \), \( T_{\text{coarse}}\left(t\right) \) the coarse-resolution reanalysis data from ERA5, \(  \rho \) the amplifier parameter,  and \( \mathcal{N}_{conv} \) the convolutional layers. They are used to model the overall temporal trend, daily fluctuations and spatiotemporal variations, respectively. In practice, we used the annual mean spectral reflectance as $X_t$. The annual temperature cycle component can be written out as 
\begin{equation}
    \mathcal{T}_{\text{ATC}}(t) = T_0 + A \sin\left( \frac{2\pi t}{N_{DoY}} + \phi \right)   \;,
\end{equation}
where, the three parameters are \( T_0 \) is the annual mean temperature, \( A \) is the amplitude of temperature fluctuation, \( \phi \) is phase shift, and \(N_{DoY}\) the number of days of the specific year. In practice, we separate the training by hour of the day into 24 parts for each year, i.e., one model for one hour (0:00 to 23:00). In this way, we maximize the information available within one year and minimize the cloud blocking effect. 

\begin{figure}[htbp]
    \centering
    \includegraphics[width=0.7\linewidth]{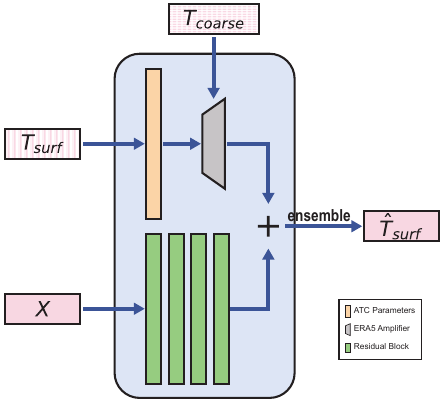}
    \caption{The Amplifier of the proposed approach.}
    \label{fig:model01}
\end{figure}

The Amplifier model takes inputs of the observed, incomplete surface temperature data  and auxiliary input of the spectral reflectance. These observed data are used to estimate the ATC parameters and the amplifier parameter related to the coarse temperature data from climate reanalysis, representing the overall temporal trend and daily fluctuations. Then, for spatiotemporal variation, the convolutional layers consist of four residual blocks~\citep{he2016deep}, each composed of \(3 \times 3\) convolutional layers (Figure~\ref{fig:model01}). These convolutional layers use the auxiliary input and gradually estimate the spatiotemporal residuals not estimated from the previous components. The first residual block increases the number of channels from 5 to 16, followed by subsequent blocks that upscale from 16 to 64, 64 to 128, and finally from 128 to 365 or 366 channels, depending on the year. These convolutional layers replace the conventional external spatiotemporal filters often used in surface temperature reconstruction~\citep{2025arXiv250214433L, zhang2021global}, enabling fully end-to-end training. As the model incorporates both the annual temperature cycle and an amplified ERA5 signal---both derived from physics-based models that encode substantial physical constraints---the Amplifier model remains grounded in physical laws while retaining the flexibility to fit observed GOES-16 surface temperature data.

\subsubsection{Transforming surface temperature to air temperature}
We use another neural network \(\mathcal{T}\) to transform the reconstructed surface temperature (\(\hat{T}_\text{surf}^{\mathcal{A}}\)) into near-surface air temperature (\(T_\text{air}\)):  
\begin{equation}
    T_\text{air} \left(t\right) = \mathcal{T} \left( \hat{T}_\text{surf}^{\mathcal{A}} \left(t\right), X_t \right) \;.
\end{equation}
In addition to the reconstructed surface temperature, we include additional variables (\(X_t\)) to capture spatiotemporal patterns, such as spatiotemporal indices (latitude, longitude, hour of the day), DEM and slope to account for elevation effects, and five hourly low-resolution ERA5 climate reanalysis variables (boundary layer height, total column water, surface sensible heat flux, zonal wind, and meridional wind), which provide crucial information on heat exchange between the Earth's surface and near-surface air. Details on these variables are described in section \ref{ssec:features}. 

\begin{figure}[htbp]
    \centering
    \includegraphics[width=0.8\linewidth]{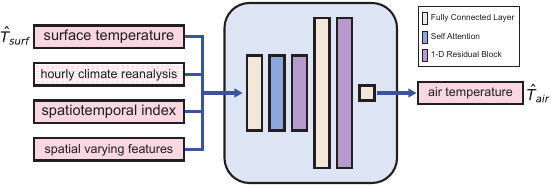}
    \caption{The Air-Transformer of the proposed approach.}
    \label{fig:model02}
\end{figure}

The second neural network of the proposed approach includes a fully connected layer with 64 neurons followed by ReLU activation, a self attention layer (64 feature maps), a residual block (64 feature maps), another fully connected layer followed by ReLU activation (128 feature maps),  another residual block (128 feature maps), and finally a fully connected layer to output the near-surface air temperature (Figure~\ref{fig:model02}). After reviewing the literature~\citep{zhang2022hourly} and conducting an initial test, we separate the training for each month. The relationship between surface temperature and air temperature varies seasonally, and implementing the model on a monthly basis helps capture these variations. As the main input is the reconstructed surface temperature and all other inputs are spatiotemporal indices and climate-related variables, the model is bounded by their latent relationship.

\subsubsection{Finding model solutions via deep ensemble learning}
We minimize the mean absolute error loss function (equivalent to maximize the Laplace likelihood) to train the two neural networks: 
\begin{equation}
\hat{\theta}_{\mathcal{A}} = \arg \min_{\theta_{\mathcal{A}} } \mathbb{E}\left[ | T_\text{surf} - \hat{T}_\text{surf}^{\mathcal{A}} | \right]  \;,
\end{equation}
\begin{equation}
\hat{\theta}_{\mathcal{T}} = \arg \min_{\theta_{\mathcal{T}} } \mathbb{E}\left[ | T_\text{air} - \hat{T}_\text{air}^{\mathcal{T}} | \right]  \;.
\end{equation}
In the Amplifier model, we obtain 200 model snapshots through the optimization trajectory using deep ensemble learning~\citep{huang2017snapshot}, achieving prediction mean: 
\begin{equation}
\hat{T}_{\text{surf}}^{\theta} = \sum_{k=1}^{200} w_k \hat{T}_{\text{surf}}^{\mathcal{A}_k}  \;,
\end{equation}
where \( \hat{T}_{surf}^{\theta} \) represents the final ensemble prediction of surface temperature, and \( w_k \) are the weights assigned to each model \( \mathcal{A}_k \) based on certain criteria. We use equal weight in this study, and this ensemble process can be interpreted as approximating the posterior predictive distribution under the assumption of a uniform prior over the model parameters.

\begin{figure}[htbp]
    \centering
    \includegraphics[width=0.7\linewidth]{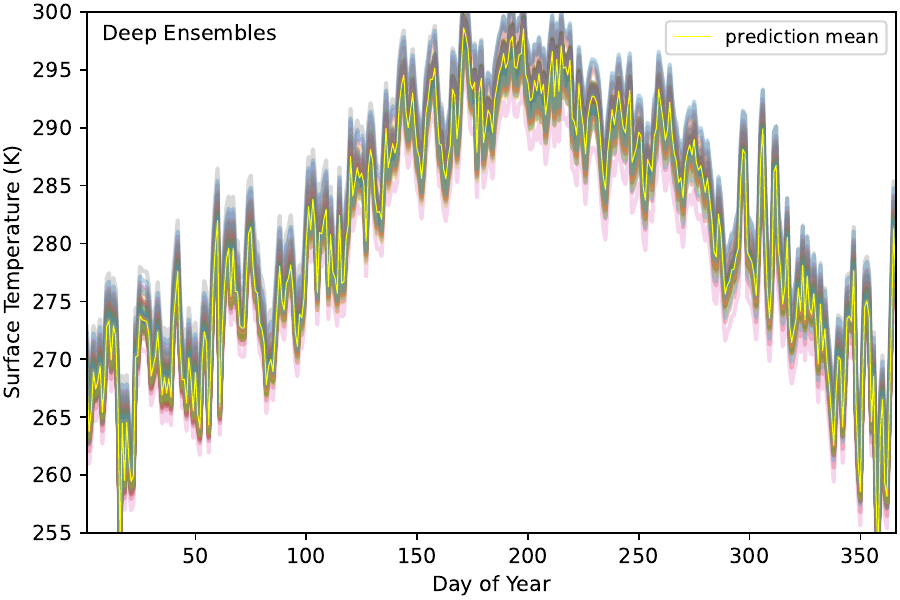}
    \caption{Deep ensembles over surface temperature prediction. All 200 predictions are shown with 50\% transparency, and the mean prediction is highlighted in yellow.}
    \label{fig:DeepEnsembleDemo}
\end{figure}

Figure~\ref{fig:DeepEnsembleDemo} presents a demonstration of deep ensemble learning. All 200 model snapshots represent potential solutions that reproduce the observed surface temperature, and as such, each is considered valid. By leveraging this ensemble, we can inherently incorporate uncertainty quantification within the model, enabling the computation of prediction intervals without significantly increasing computational complexity.

\subsubsection{Non-parametric uncertainty estimation}
Given an ensemble of 200 models producing predictions \( \hat{T}_{surf}^{(\mathcal{A}_1)}, \hat{T}_{surf}^{(\mathcal{A}_2)}, \dots, \hat{T}_{surf}^{(\mathcal{A}_{200})} \) for each data point, we estimate uncertainty based on the variability among these predictions. The underlying principle is that near the end of training, a neural network can converge to multiple possible solutions within the loss landscape. Data points exhibiting greater variability in predictions inherently correspond to higher uncertainty, serving as an indicator of model confidence.
From the perspective of parameter space, each snapshot in the ensemble represents a valid solution near a local minimum. By aggregating these snapshots, the ensemble provides a more comprehensive characterization of the loss landscape's structure, capturing a broader range of possible model solutions. To quantify uncertainty, we construct an initial prediction interval using the 2.5\% and 97.5\% percentiles. Specifically, the predictions from all 200 models are first sorted in ascending order:
\begin{equation}
\hat{T}_{surf}^{(1)} \leq \hat{T}_{surf}^{(2)}  \leq \dots \leq \hat{T}_{surf}^{(200)}  \;.
\end{equation}
We calibrate the uncertainty quantification by ensuring that the prediction intervals cover 95\% of the training data using a calibration factor \( \lambda \). Let \( d_L = \hat{T}_{\text{surf}}^{\theta} - \hat{T}_{\text{surf}}^{(5)} \) and \( d_U = \hat{T}_{\text{surf}}^{(195)} - \hat{T}_{\text{surf}}^{\theta} \). The final prediction interval is calibrated over the training data:
\begin{equation}
\left[ \hat{T}_{\text{surf}}^{\theta} - \lambda d_L, \hat{T}_{\text{surf}}^{\theta} + \lambda d_U \right] \;.
\end{equation}

This interval captures the spread of the central 95\% of the model predictions, providing a non-parametric uncertainty estimate based on the disagreement among the 200 model solutions. Using a non-parametric uncertainty estimate offers flexibility and robustness, particularly in capturing extreme temperature events, as it does not assume a specific underlying distribution for the data. This is especially useful given the complexity of satellite and Earth data~\citep{cardellach2008new}.

Finally, we propagate the uncertainty estimates from the surface temperature prediction to near-surface air temperature, with the upper and lower predictions as
\begin{equation}
    \left[ \hat{T}_\text{air}^{\text{low}} \left(t\right), \hat{T}_\text{air}^{\text{upp}} \left(t\right) \right] =   \left[ \mathcal{T} \left( \hat{T}_{\text{surf}}^{\theta} - \lambda d_L, X_t \right), \mathcal{T} \left( \hat{T}_{\text{surf}}^{\theta} + \lambda d_L, X_t \right) \right] \;.
\end{equation}
We assume the uncertainty of the air temperature estimation can be reflected through surface temperature data, which are reconstructed surface temperature data that is the output of the Amplifier module. Since the surface temperature predictions are constrained by a physics-guided neural network, both the predictions and the associated uncertainty estimates remain linked to the reconstructed surface temperature. This ensures that all predictions are consistent with physical laws and aligned with observations, thereby maintaining the reliability and accuracy of the final model outputs.

\section{Results and Analysis}
\label{sec:ResultAnalysis}

\subsection{Implementation details}
We develop and test the approach in Python 3.9 using the PyTorch deep learning framework~\citep{10.5555/3454287.3455008}. The Amplifier model is developed as a single convolutional neural network model encoded with the annual temperature cycle. In training the models, we divide the study area into six parts and train the Amplifier model first, separately for each hour each year, using the Adam optimizer with a learning rate of 0.1. We train the Amplifier model for 600 epochs and obtain 200 snapshots every 2 epochs after the 200th epoch when the training loss is sufficiently small. After obtaining reconstructed surface temperature, including mean and prediction intervals, we then train the second neural network separately for each month each year, using the Adam optimizer with a learning rate of 0.01 and a batch size of 65,536 for 500 epochs. In this part, 20\% of the meteorological stations are held-out for testing. The computation is accelerated using a GPU. All experiments are conducted on a workstation equipped with an Intel E5-2680 v4 CPU, an NVIDIA GTX 1080 Ti GPU, and 128 GB RAM.

\subsubsection{Validation metrics}
We use  root mean square error (RMSE),  mean absolute error (MAE), and \( R^2 \) to evaluate the results on the 20\% of stations not used in training: 
\begin{equation}
    \text{RMSE} = \sqrt{\frac{1}{N} \sum_{i=1}^{N} \left( T_{\text{air}, i} - \hat{T}_{\text{air}, i} \right)^2}  \;, 
\end{equation}
\begin{equation}
    \text{MAE} = \frac{1}{N} \sum_{i=1}^{N} \left| T_{\text{air}, i} - \hat{T}_{\text{air}, i} \right|  \;, 
\end{equation}
\begin{equation}
    R^2 = 1 - \frac{\sum_{i=1}^{N} \left( T_{\text{air}, i} - \hat{T}_{\text{air}, i} \right)^2}{\sum_{i=1}^{N} \left( T_{\text{air}, i} - \bar{T}_{\text{air}} \right)^2}  \;.
\end{equation}

\subsection{Main results}
\subsubsection{Hourly analysis}
\begin{figure*}[htbp]
    \centering
    \includegraphics[width=1.0\linewidth]{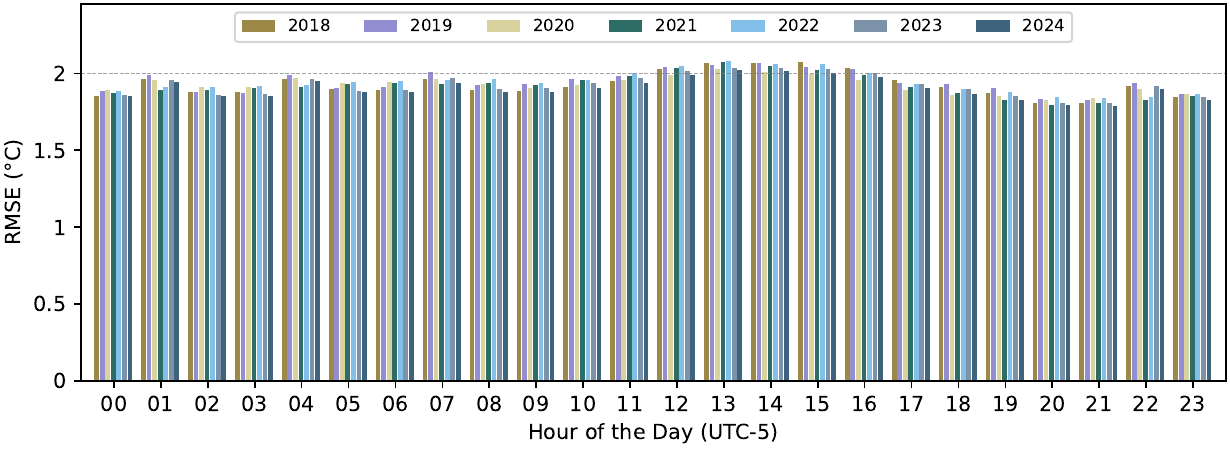}
    \caption{Hourly RMSE of near-surface air temperature estimation for each year, with the hour of the day set to UTC-5 (Eastern Time).}
    \label{fig:rmseHour}
\end{figure*}

The proposed approach achieved accurate estimations of hourly air temperature, with an overall RMSE of 1.93\degree C. Figure~\ref{fig:rmseHour} presents the RMSE by hour (UTC-5, Eastern Time) for each year, revealing a consistent pattern. RMSE values slightly exceed 2\degree C during midday hours (12:00--15:00), when the difference between surface and air temperature is typically larger. At other hours, the RMSE remains below 2\degree C. The RMSE ranges from 1.73\degree C to 2.13\degree C.

\subsubsection{Monthly analysis}
\begin{figure}[htbp]
    \centering
    \includegraphics[width=1.0\linewidth]{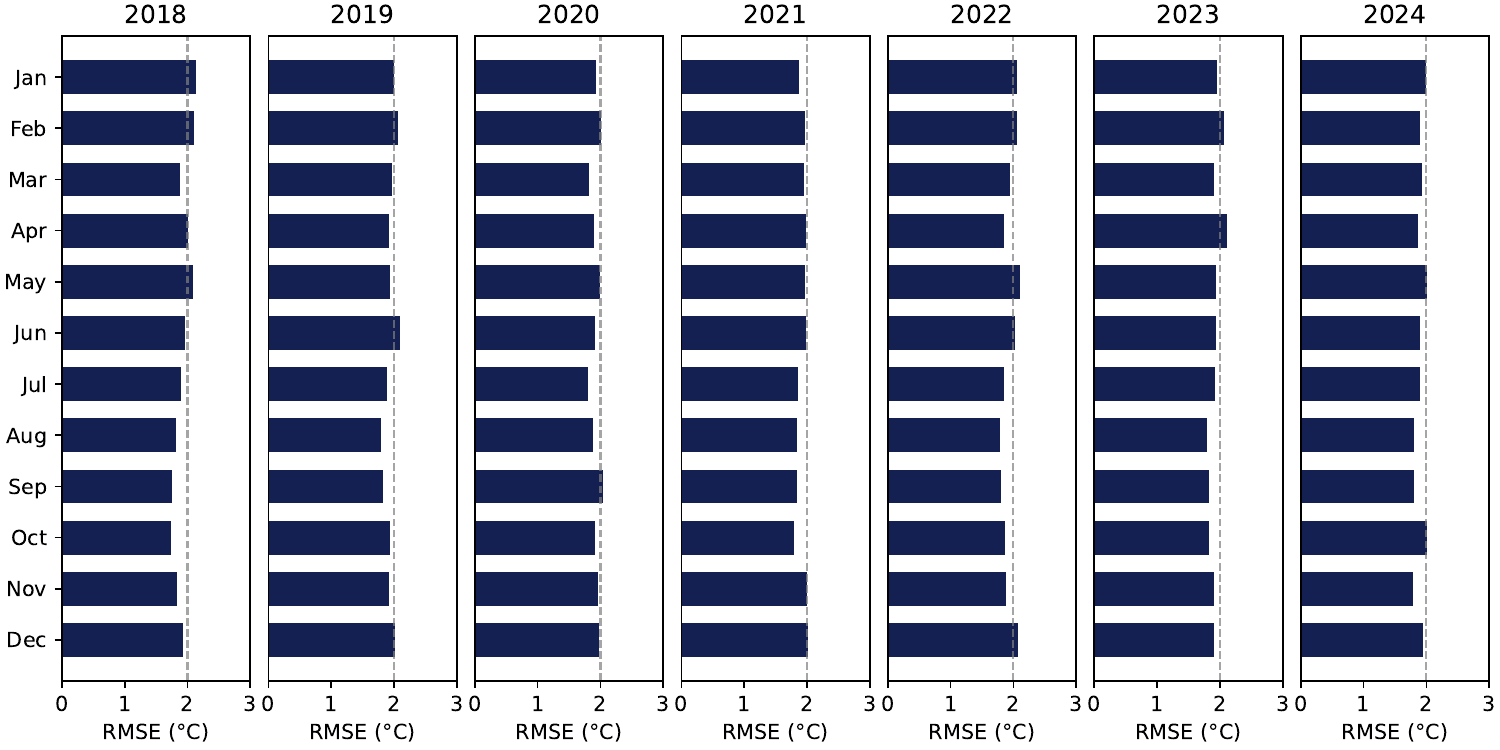}
    \caption{Monthly root mean square error (RMSE) of near-surface air temperature predictions by year.}
    \label{fig:result01}
\end{figure}

We analyze the estimation error for each month in Figure~\ref{fig:result01}. The model performs better in late summer and early fall (July, August, September, and October), and slightly worse in January, regardless of the year. This performance surpasses existing approaches for hourly estimations and is on par with other methods generated LST data from valid observations only. 

\begin{figure*}[htbp]
    \centering
    \includegraphics[width=1.0\linewidth]{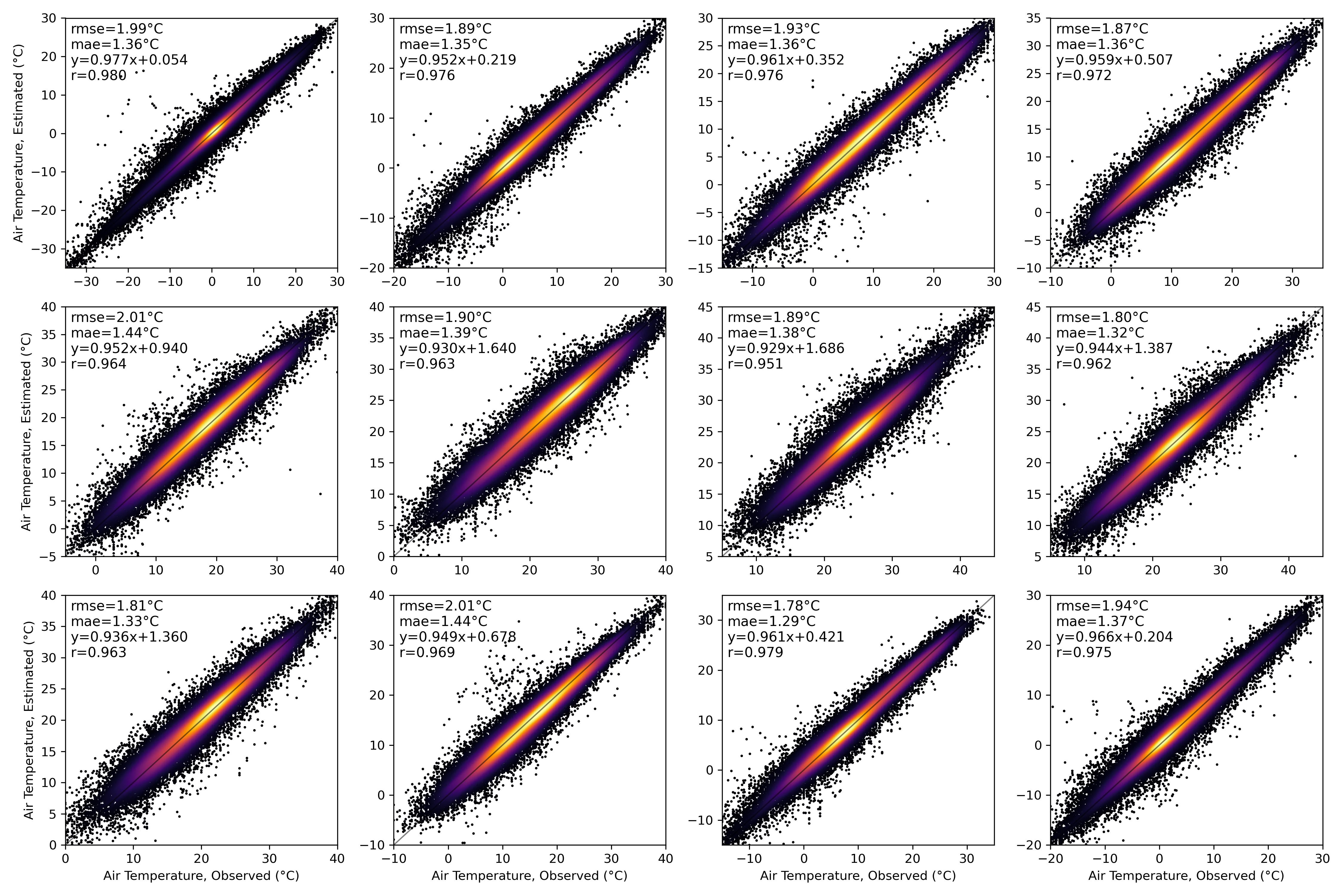}
    \caption{Scatter plot of 2024 data showing in situ vs. estimated near-surface air temperature.}
    \label{fig:scatter2024}
\end{figure*}

We present the monthly scatter plots for 2024 in Figure~\ref{fig:scatter2024}, with the x-axis representing in situ air temperature and the y-axis representing the estimated air temperature. Despite the wide temperature range (mostly between -30\degree C and 45\degree C), the proposed approach demonstrates consistently strong performance in near-surface temperature estimation, with RMSE ranging from 1.78\degree C to 2.01\degree C and MAE from 1.29\degree C to 1.44\degree C. The coefficients of the linear regressions are all slightly below 1.0, suggesting that the approach tends to be slightly conservative—favoring accurate and physically plausible temperature estimates.

\subsubsection{Sensitivity analysis}
\begin{figure*}[htbp]
    \centering
    \includegraphics[width=0.98\linewidth]{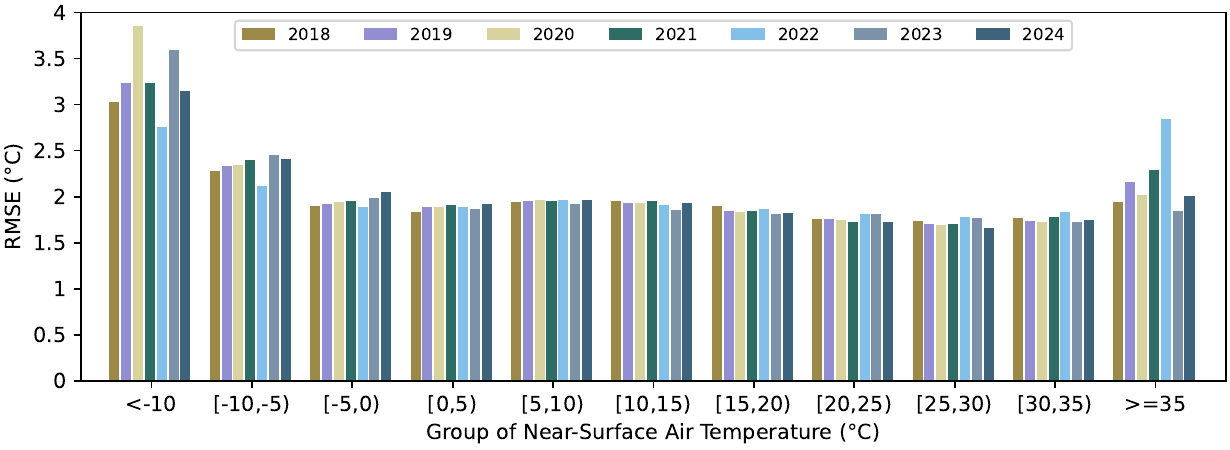}
    \caption{RMSE as a function of measured near-surface air temperature.}
    \label{fig:SenAnalysis01}
\end{figure*}

We analyzed the RMSE across the range of in-situ air temperature, as shown in Figure~\ref{fig:SenAnalysis01}. The RMSE remains relatively stable between -10\degree C and above 35\degree C. For in-situ air temperature below -10\degree C, the RMSE increases slightly by 1\degree C.

\begin{figure*}[htbp]
    \centering
    \includegraphics[width=0.24\linewidth]{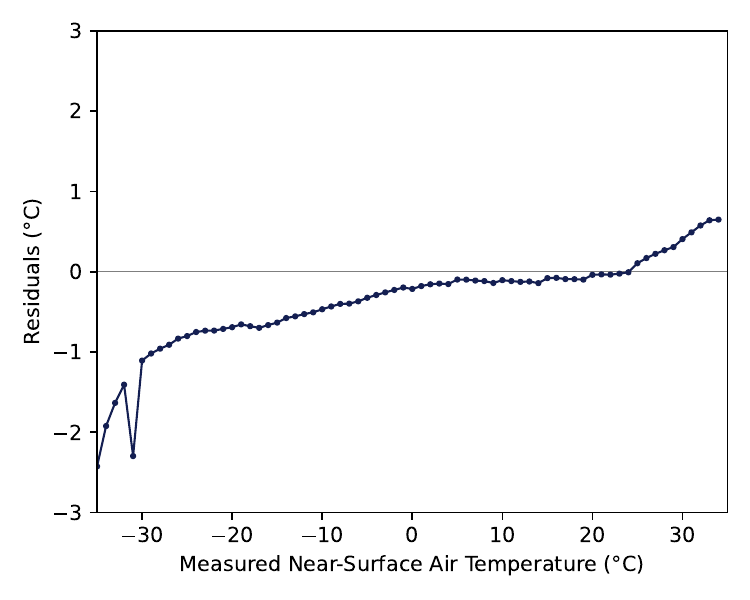}
    \includegraphics[width=0.24\linewidth]{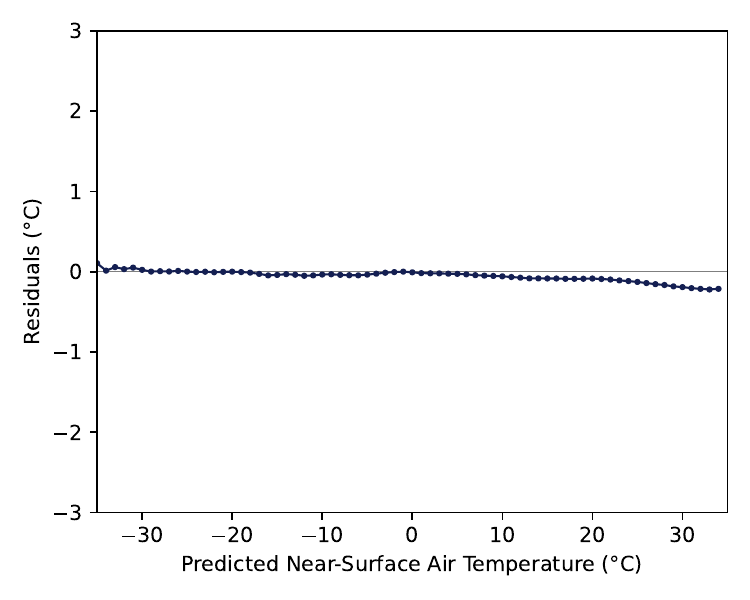}
    \includegraphics[width=0.24\linewidth]{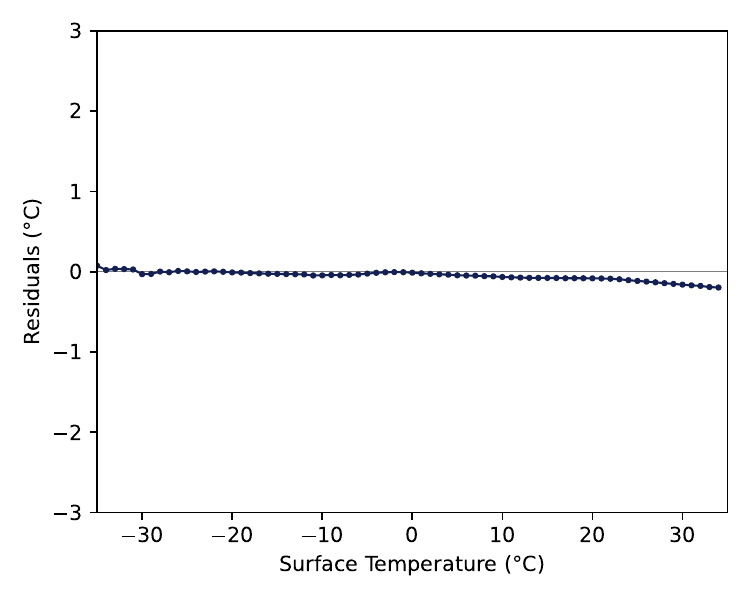}
    \includegraphics[width=0.24\linewidth]{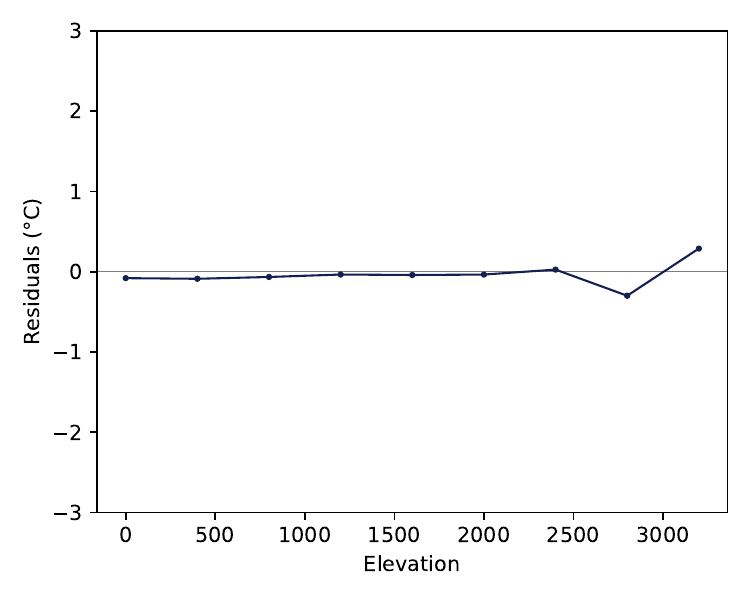}
    \caption{Residuals as a function of in situ near-surface air temperature, predicted near-surface air temperature, reconstructed surface temperature, and elevation.}
    \label{fig:SenAnalysis02}
\end{figure*}

We further examined the mean residuals and observed up to 0.5\degree C residuals when in-situ air temperature exceeds 35\degree C, and up to 1\degree C residuals when in-situ air temperature is below -10\degree C (Figure~\ref{fig:SenAnalysis02}a). This suggests that the proposed approach, consistent with previous scatter plots (Figure~\ref{fig:scatter2024}), tends to be conservative. We verified that this pattern does not apply to the residuals as a function of estimated air temperature (Figure~\ref{fig:SenAnalysis02}b), indicating that the approach is unbiased during training and that no information in the training data adjusts these residuals. For references, we show the residuals as a function of surface temperature and elevation; neither exhibits a residual pattern (Figure~\ref{fig:SenAnalysis02} [c, d]). In summary, the proposed approach successfully estimates near-surface air temperature with an overall RMSE of 1.93\degree C. The estimations are slightly conservative, underestimating air temperature by 0.5\degree C when the air temperature is higher than 35\degree C and overestimating it by 1.0\degree C when the air temperature is below -30\degree C. It is important to note that such extreme temperatures account for only a very small fraction of remote and rural areas. 

\subsection{The generated air temperature prediction maps}
\begin{figure*}[htbp]
    \centering
    \includegraphics[width=0.44\linewidth]{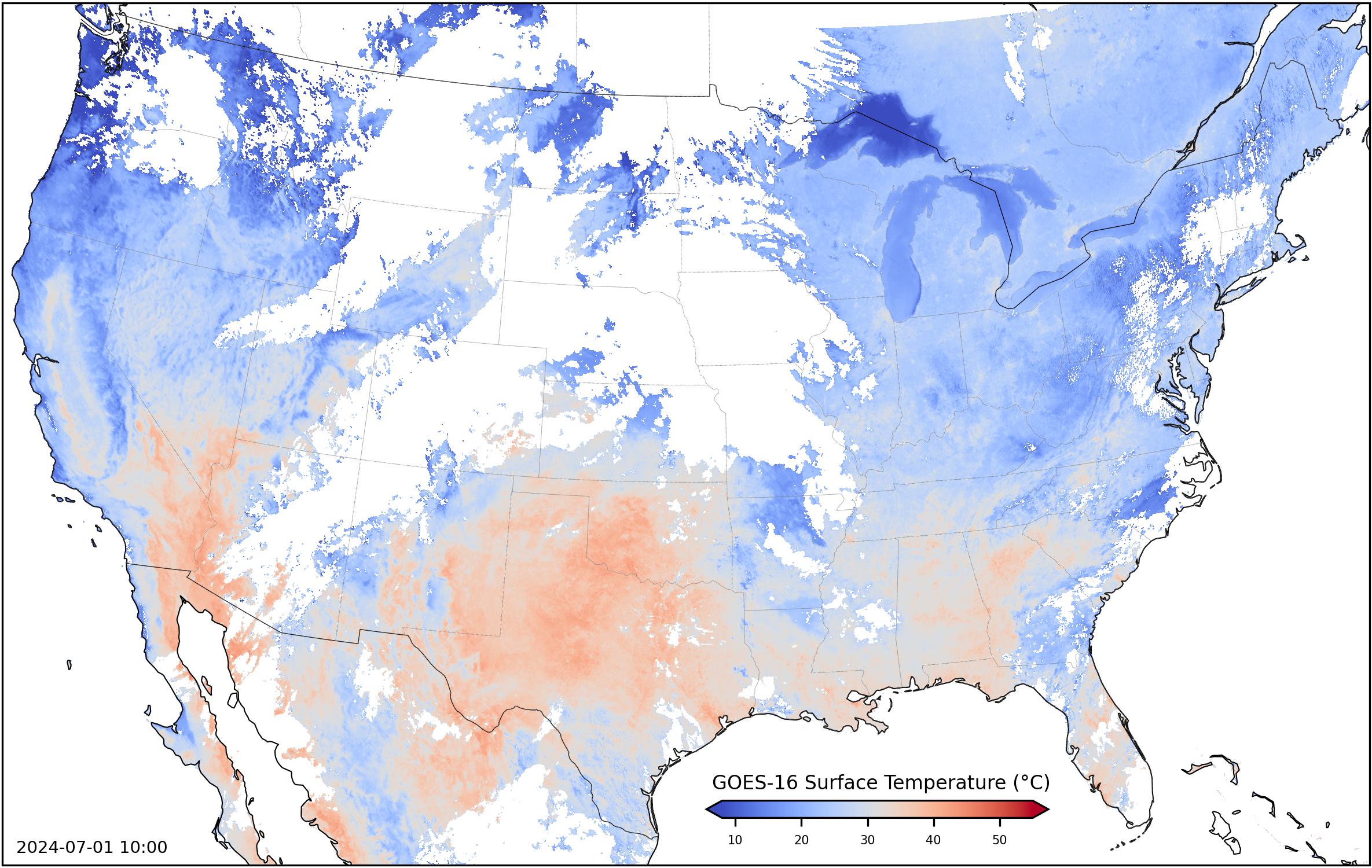}
    \includegraphics[width=0.44\linewidth]{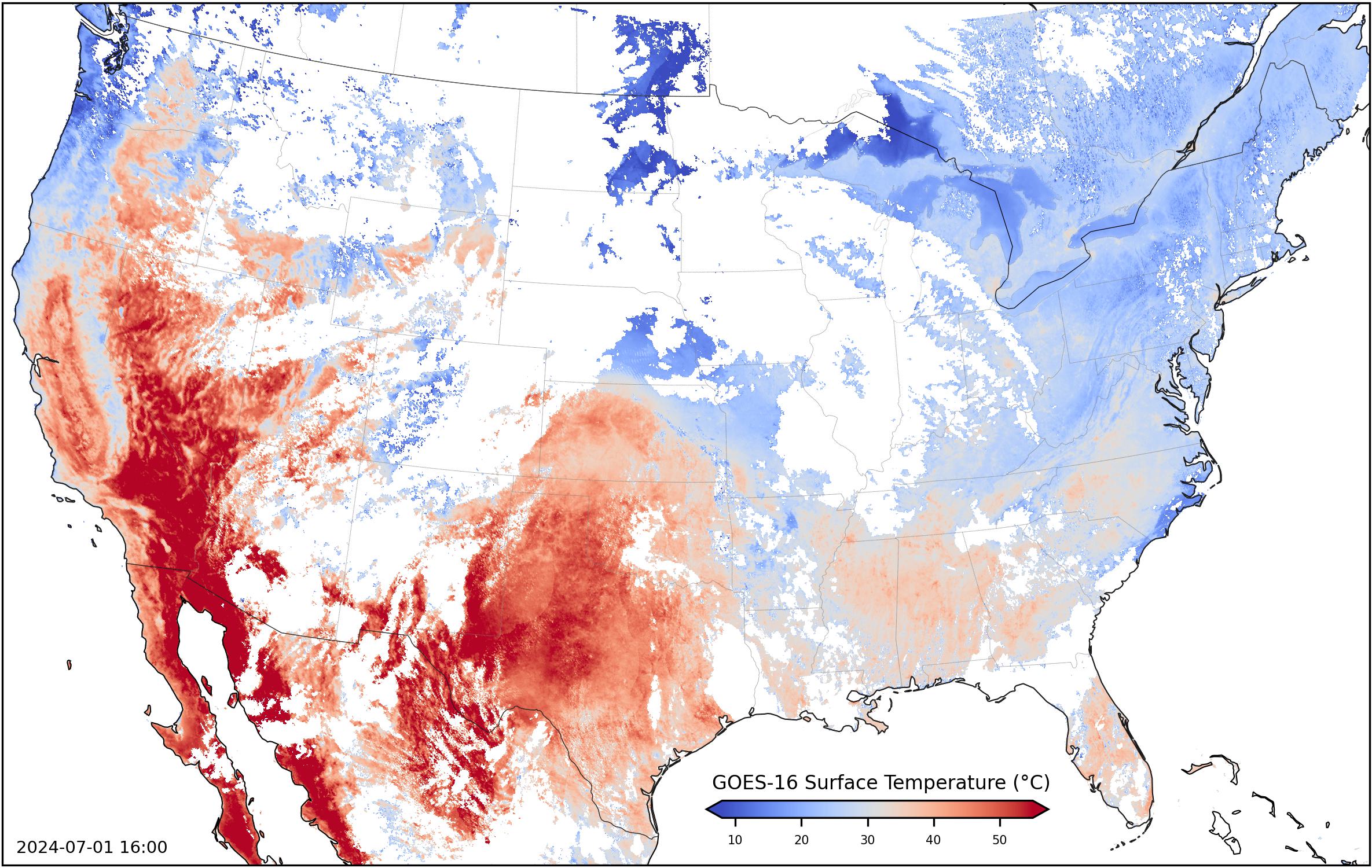} \\
    \includegraphics[width=0.44\linewidth]{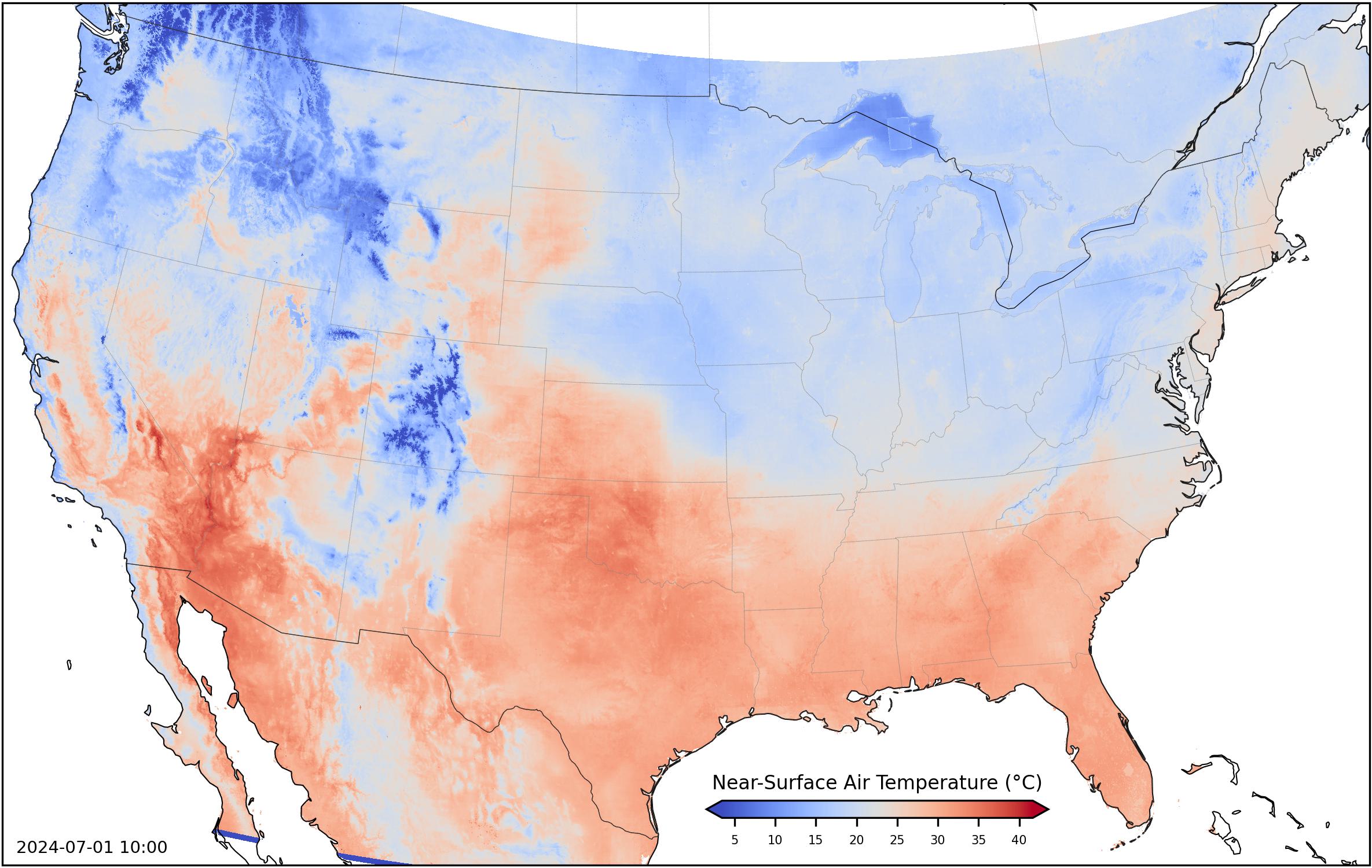}
    \includegraphics[width=0.44\linewidth]{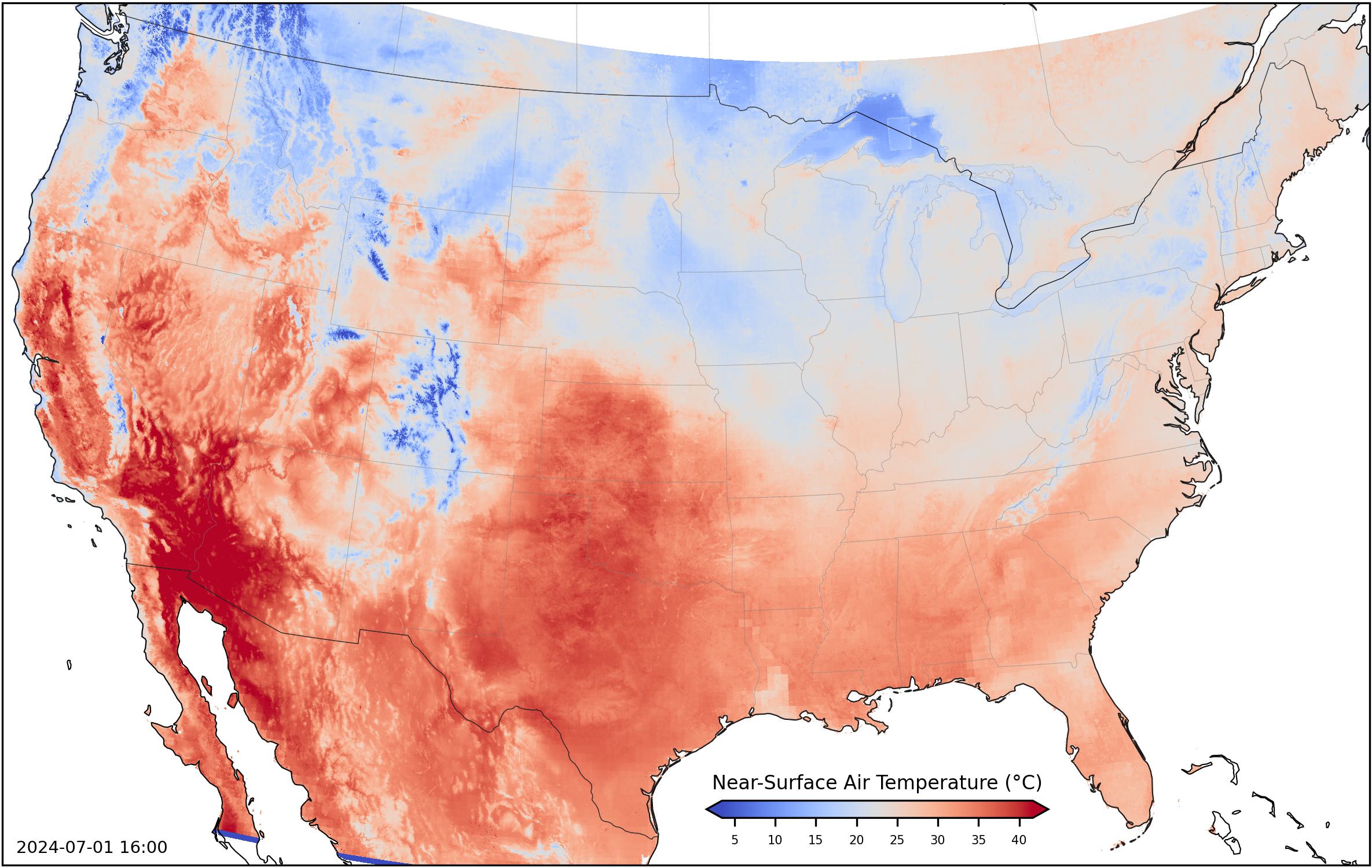} \\
    \includegraphics[width=0.44\linewidth]{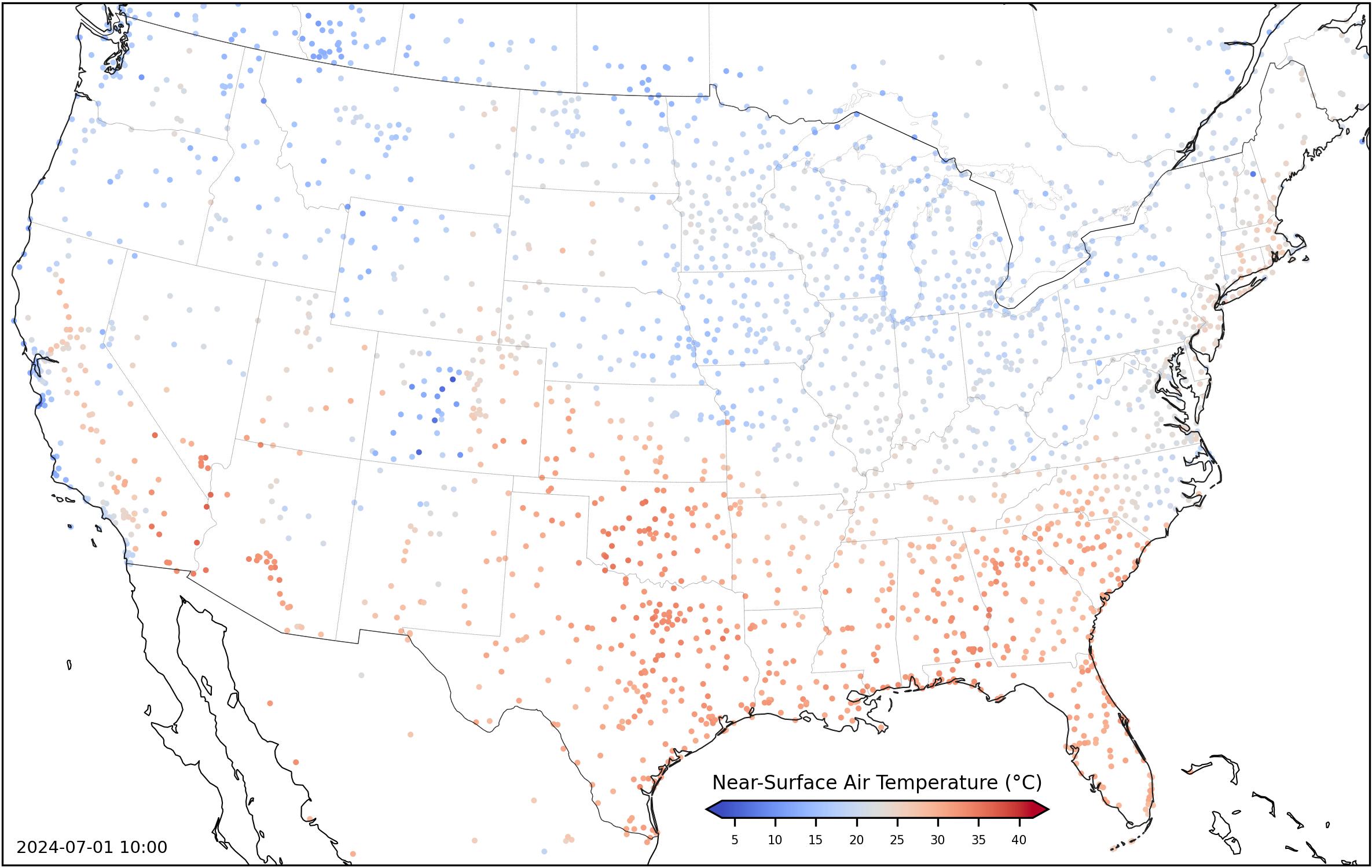}
    \includegraphics[width=0.44\linewidth]{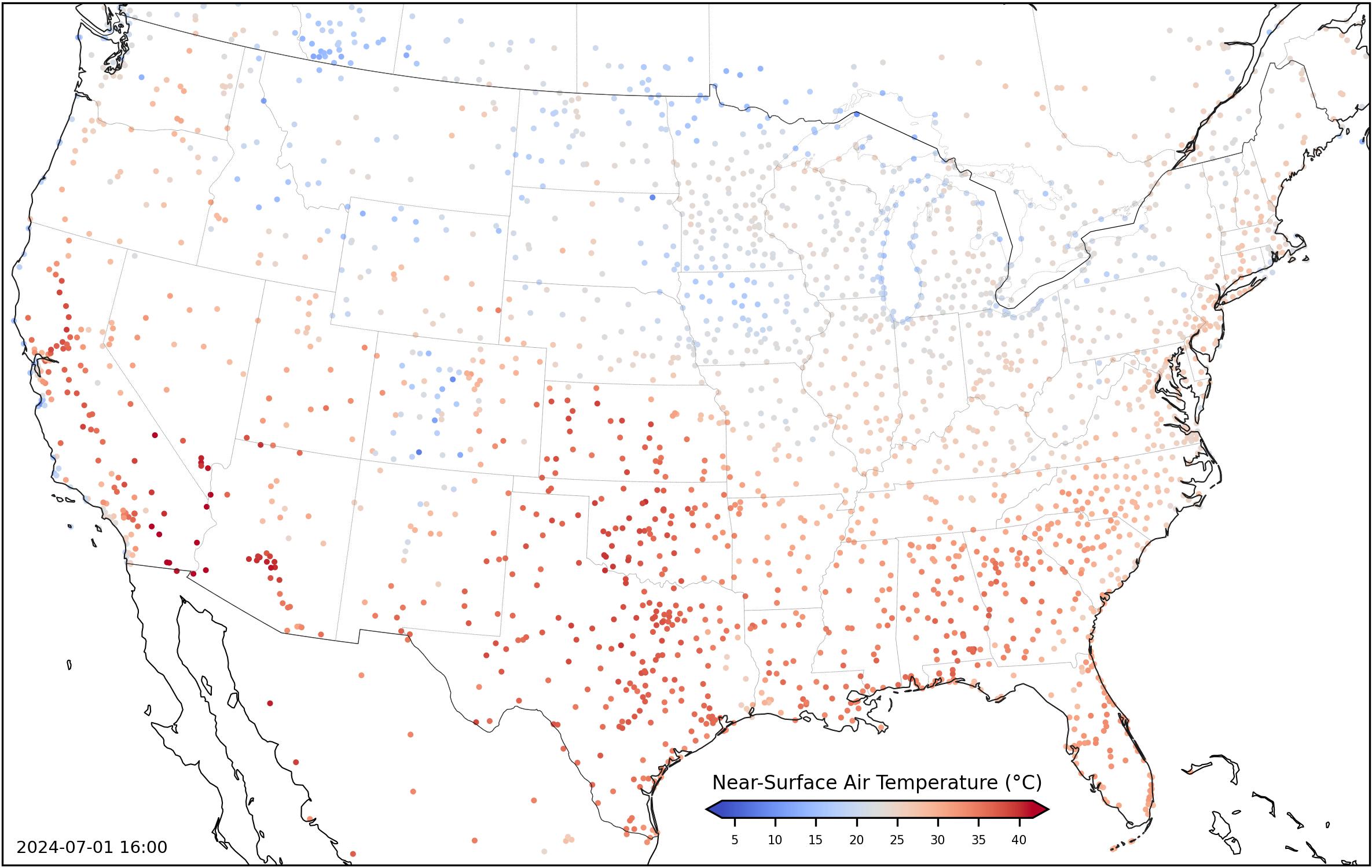} \\
    \includegraphics[width=0.44\linewidth]{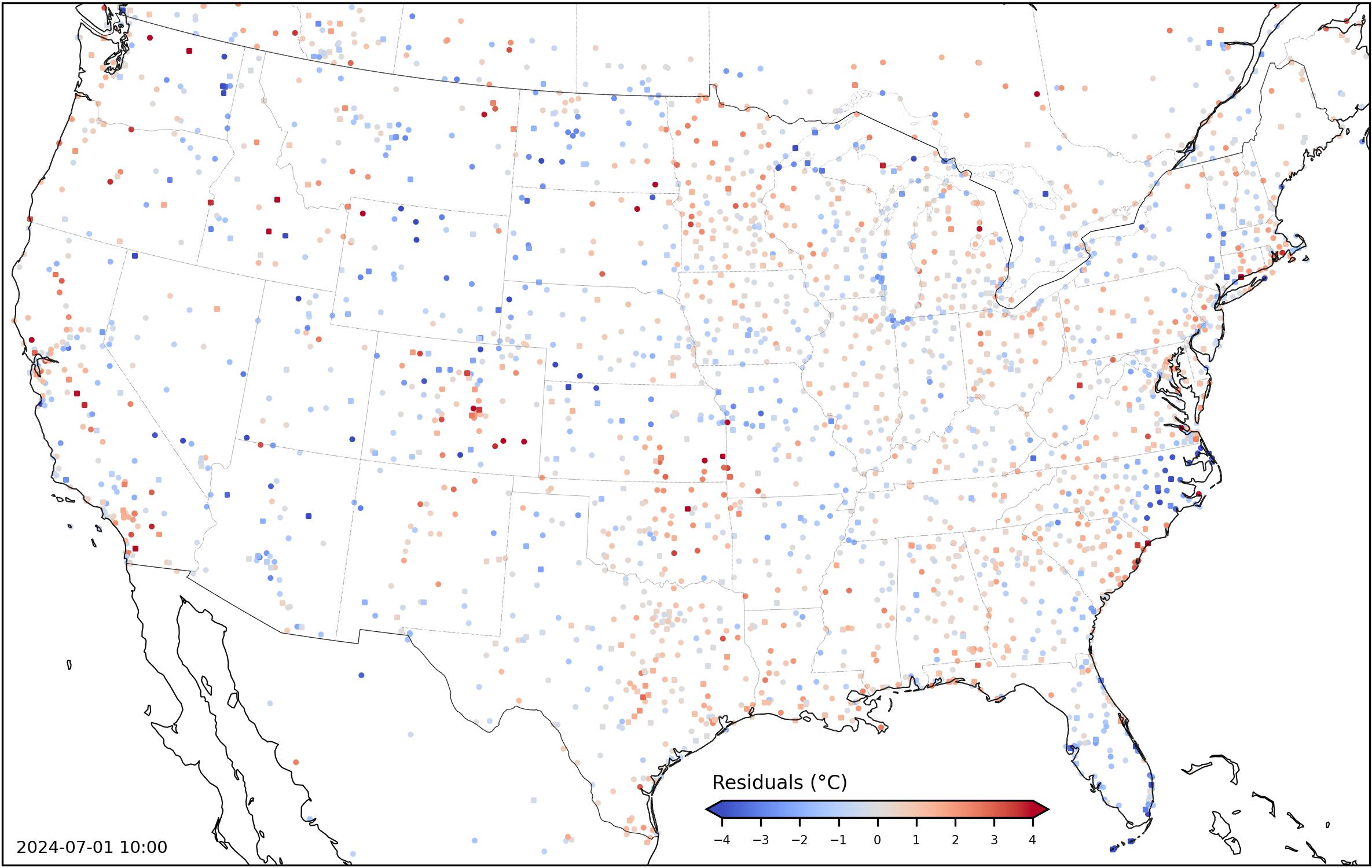} 
    \includegraphics[width=0.44\linewidth]{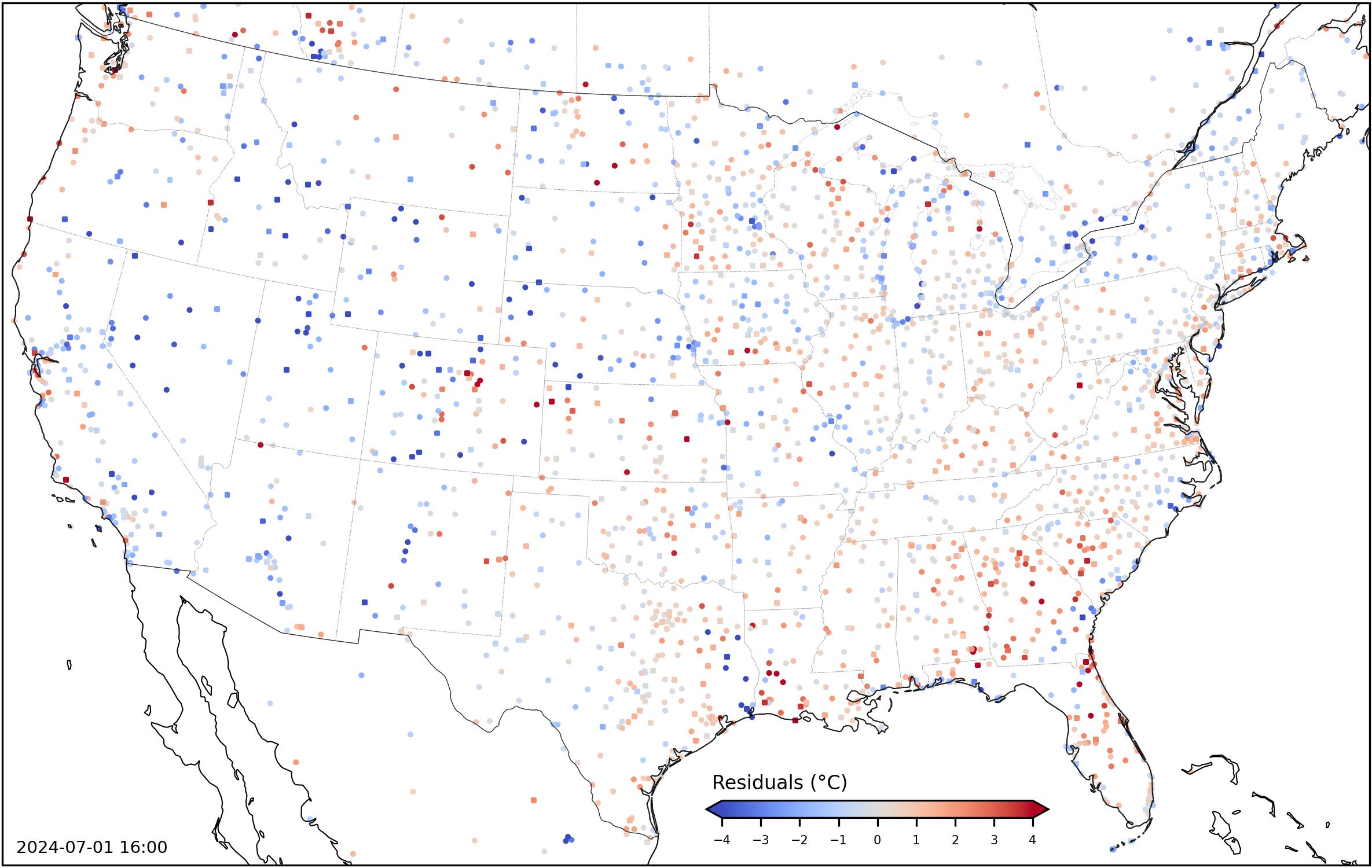}
    \caption{The cloud-obscured GOES-16 surface temperature data, the generated air temperature data, the in situ air temperature from meteorological stations, and the residuals at each station for 10:00 and 16:00 on July 1, 2024. Stations plotted as circles were used for training, and stations plotted as squares were used for testing.}
    \label{fig:maps2024a}
\end{figure*}

Here, we show two instances of the estimates on July 1, 2024, at 10:00 and 16:00 (UTC-5). We first show the cloud-obscured GOES-16 surface temperature data, the generated near-surface air temperature data, the in situ air temperature from meteorological stations, and the residuals at the station level in Figure~\ref{fig:maps2024a}. The generated air temperature data show strong agreement with the in situ measurements, exhibiting small residuals across the study area. These data successfully capture temperature variations over short periods, preserving typical patterns such as the low values in mountainous areas, the slightly cooler temperatures during the daytime over the Great Lakes, and the north-south temperature gradient. Overall, the generated air temperature data demonstrate consistent and expected spatiotemporal patterns, accurately reflecting both short-term and spatial temperature dynamics.

\begin{figure*}[htbp]
    \centering
    \includegraphics[width=0.44\linewidth]{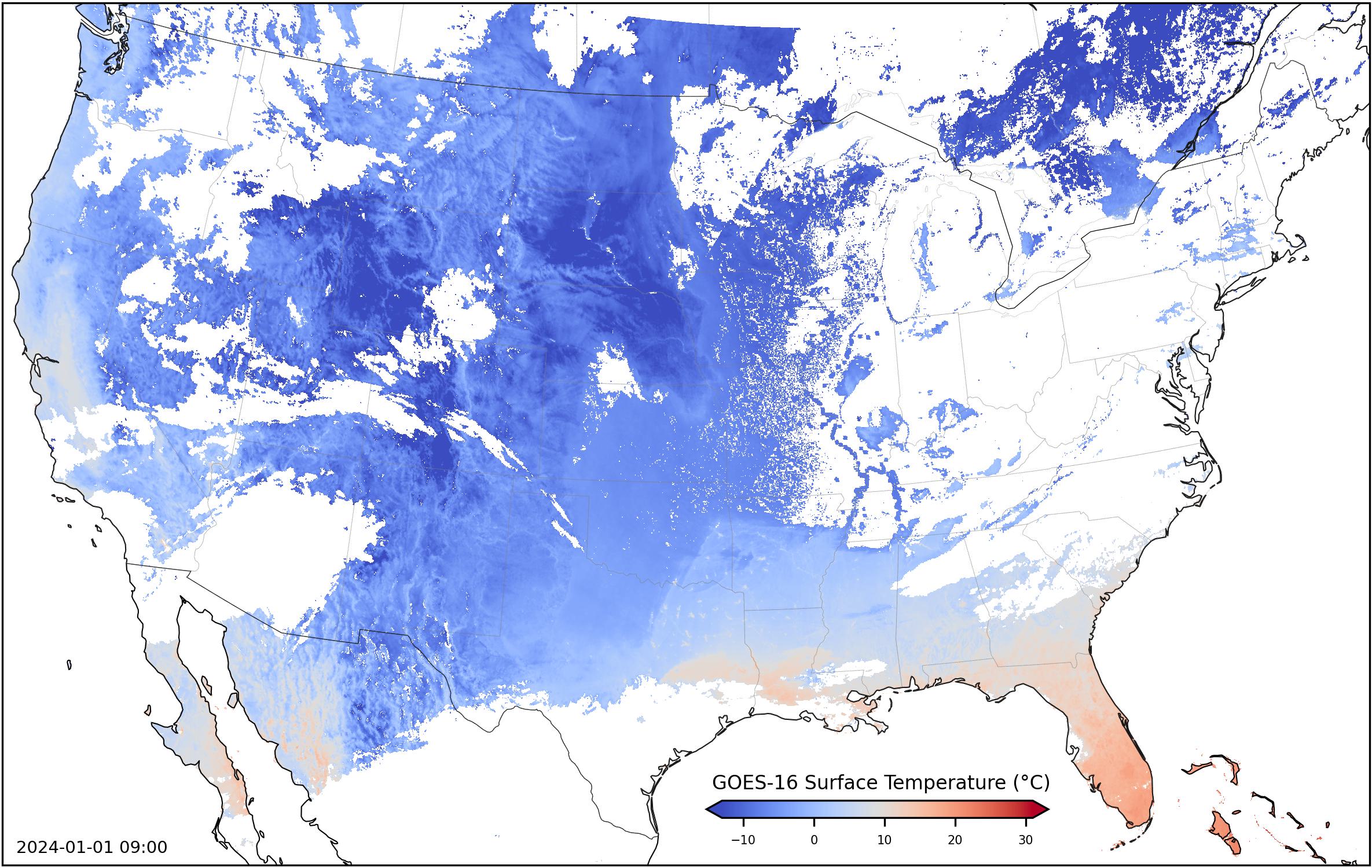}
    \includegraphics[width=0.44\linewidth]{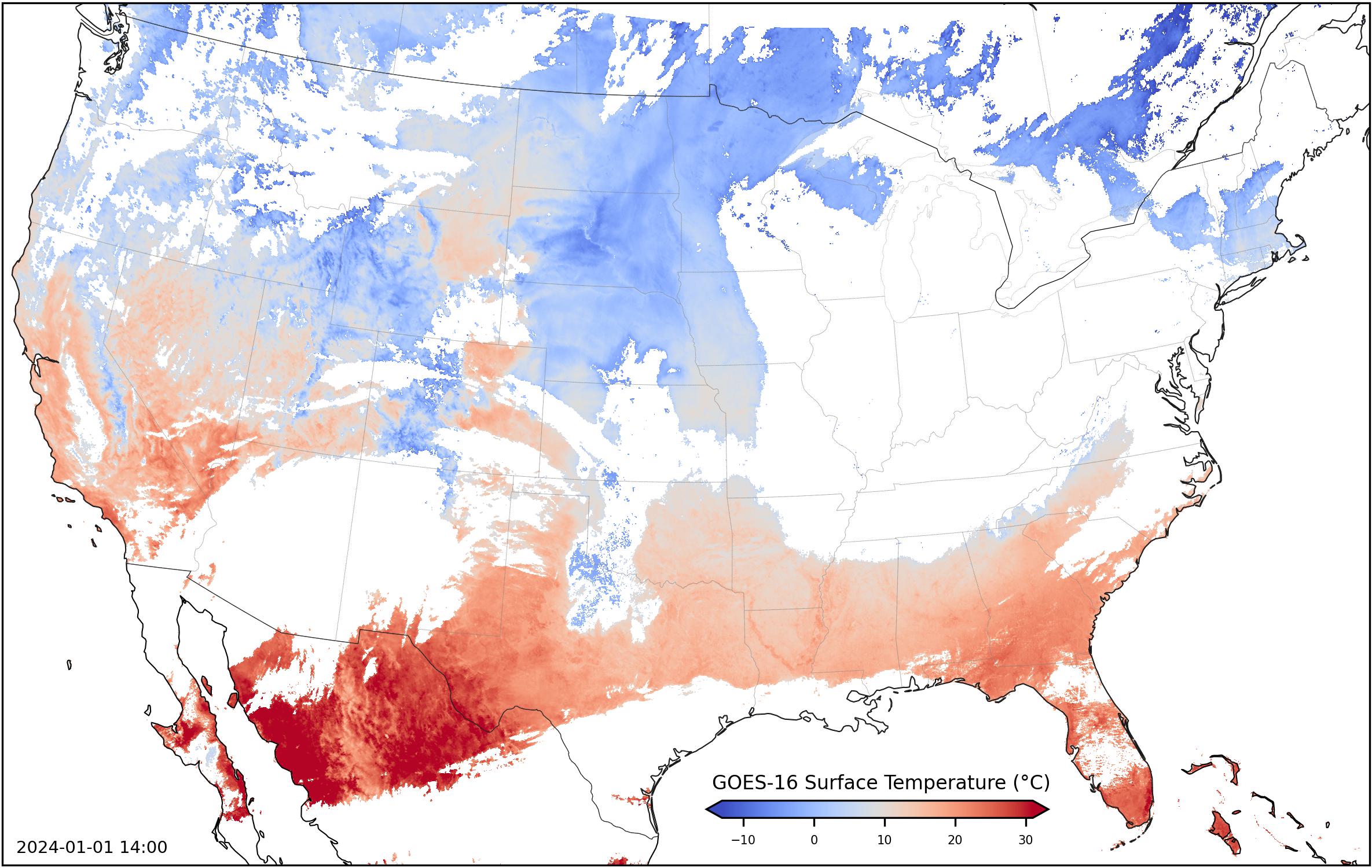} \\
    \includegraphics[width=0.44\linewidth]{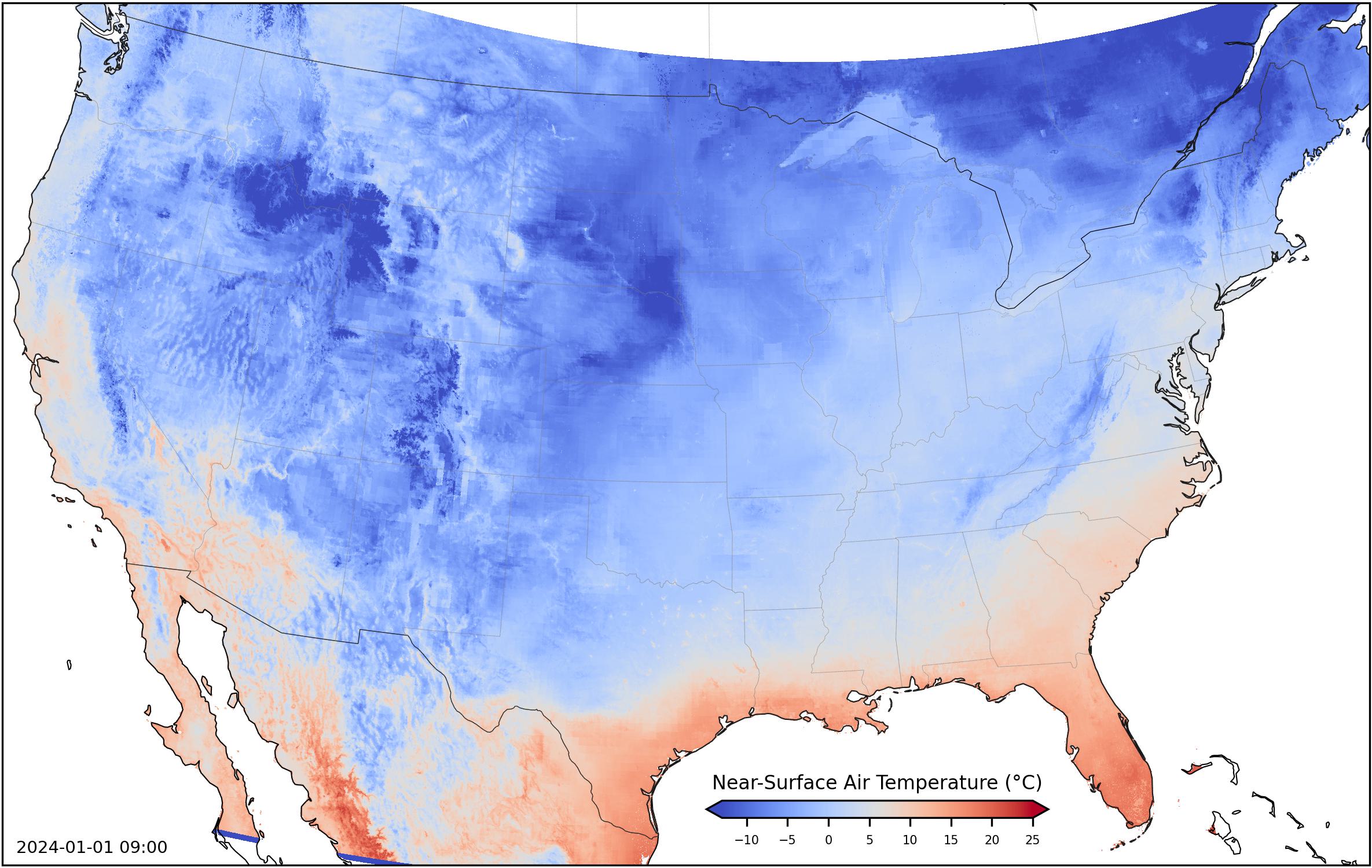}
    \includegraphics[width=0.44\linewidth]{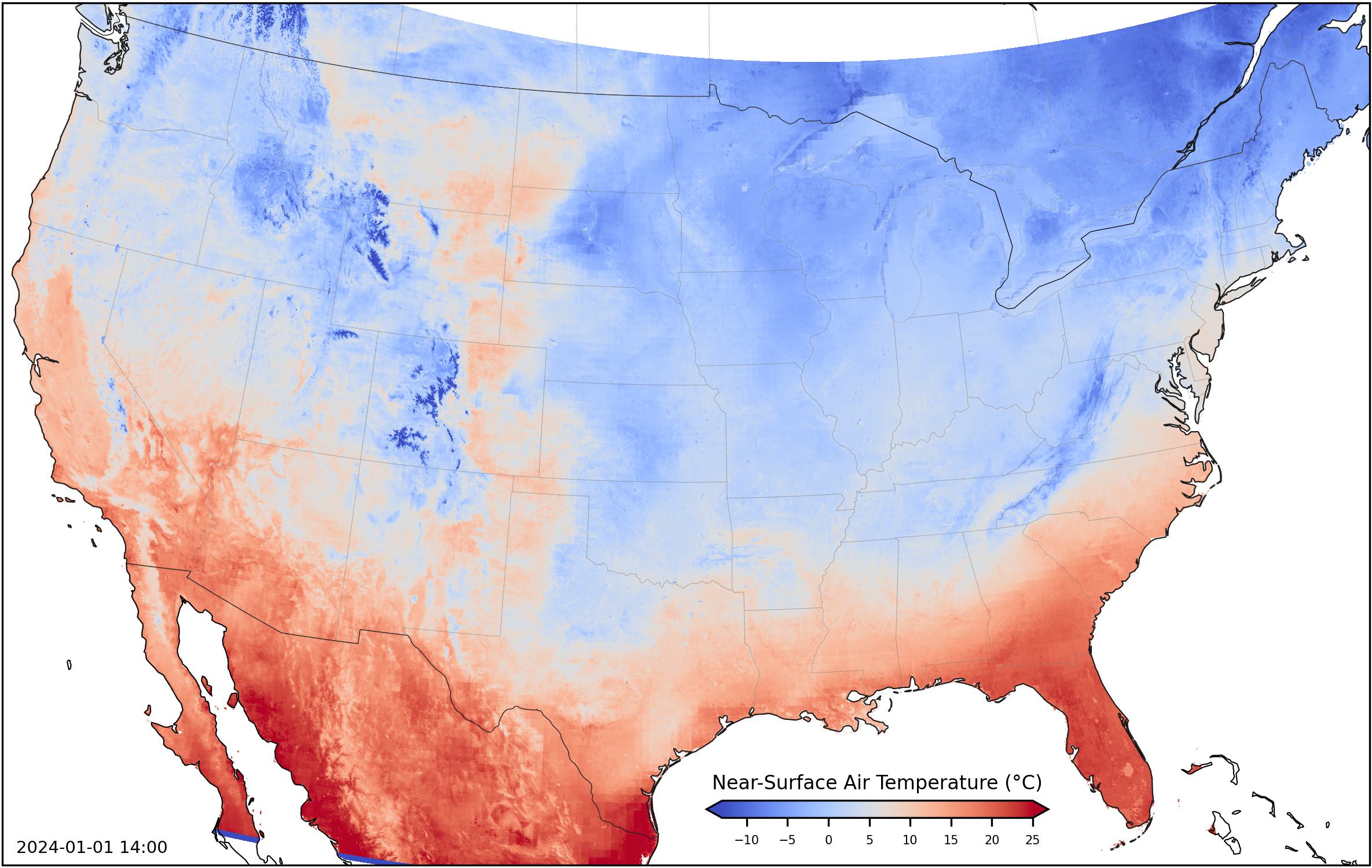} \\
    \includegraphics[width=0.44\linewidth]{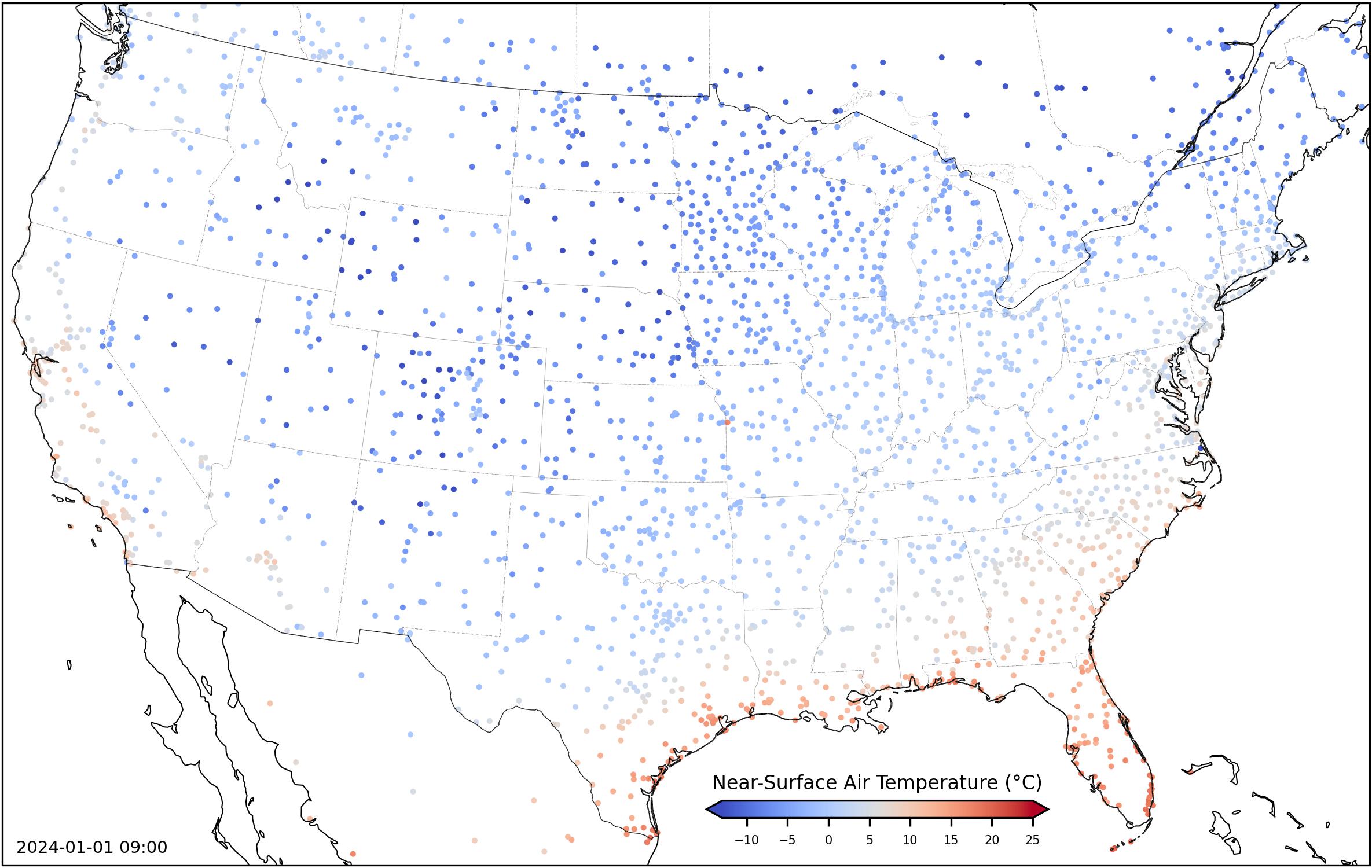}
    \includegraphics[width=0.44\linewidth]{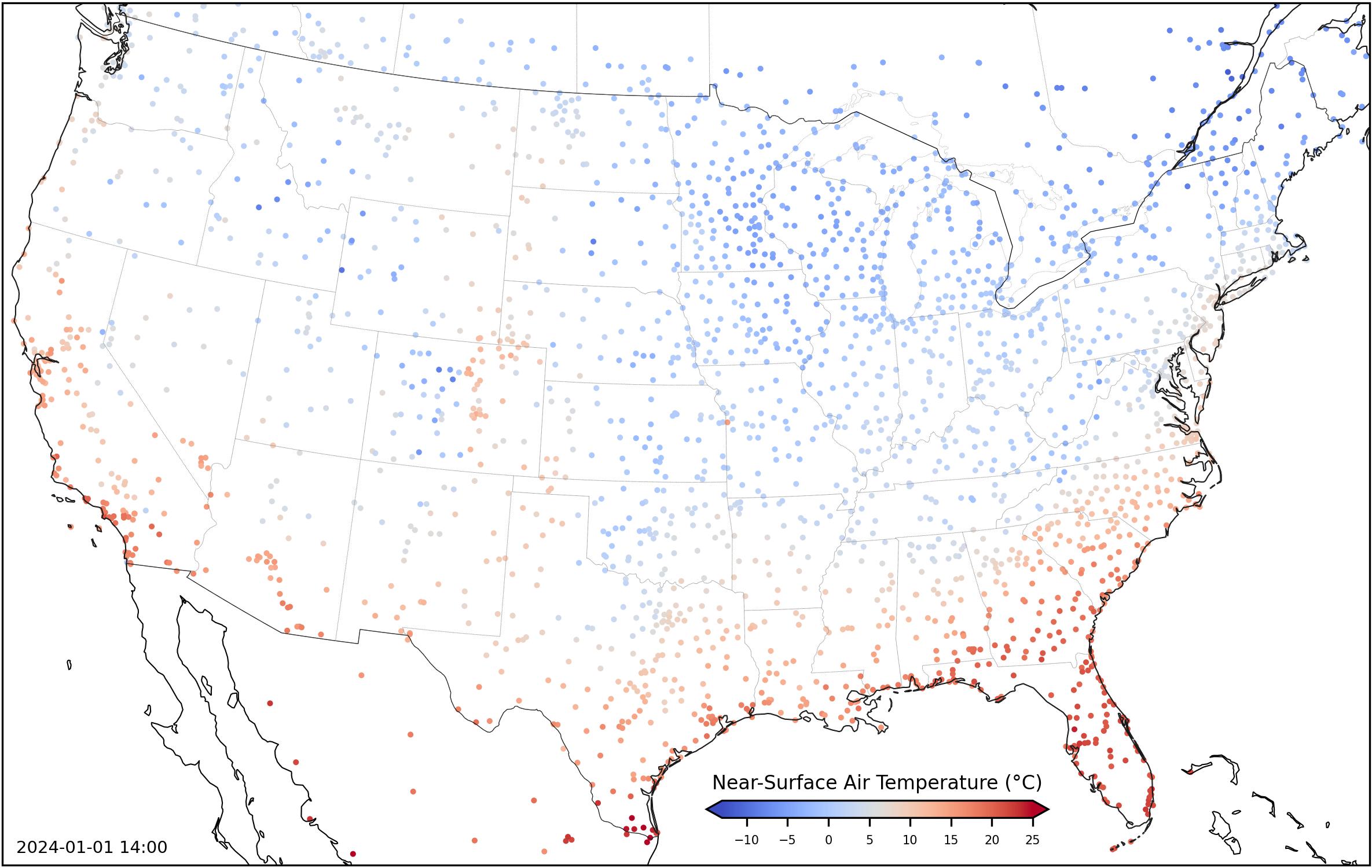} \\
    \includegraphics[width=0.44\linewidth]{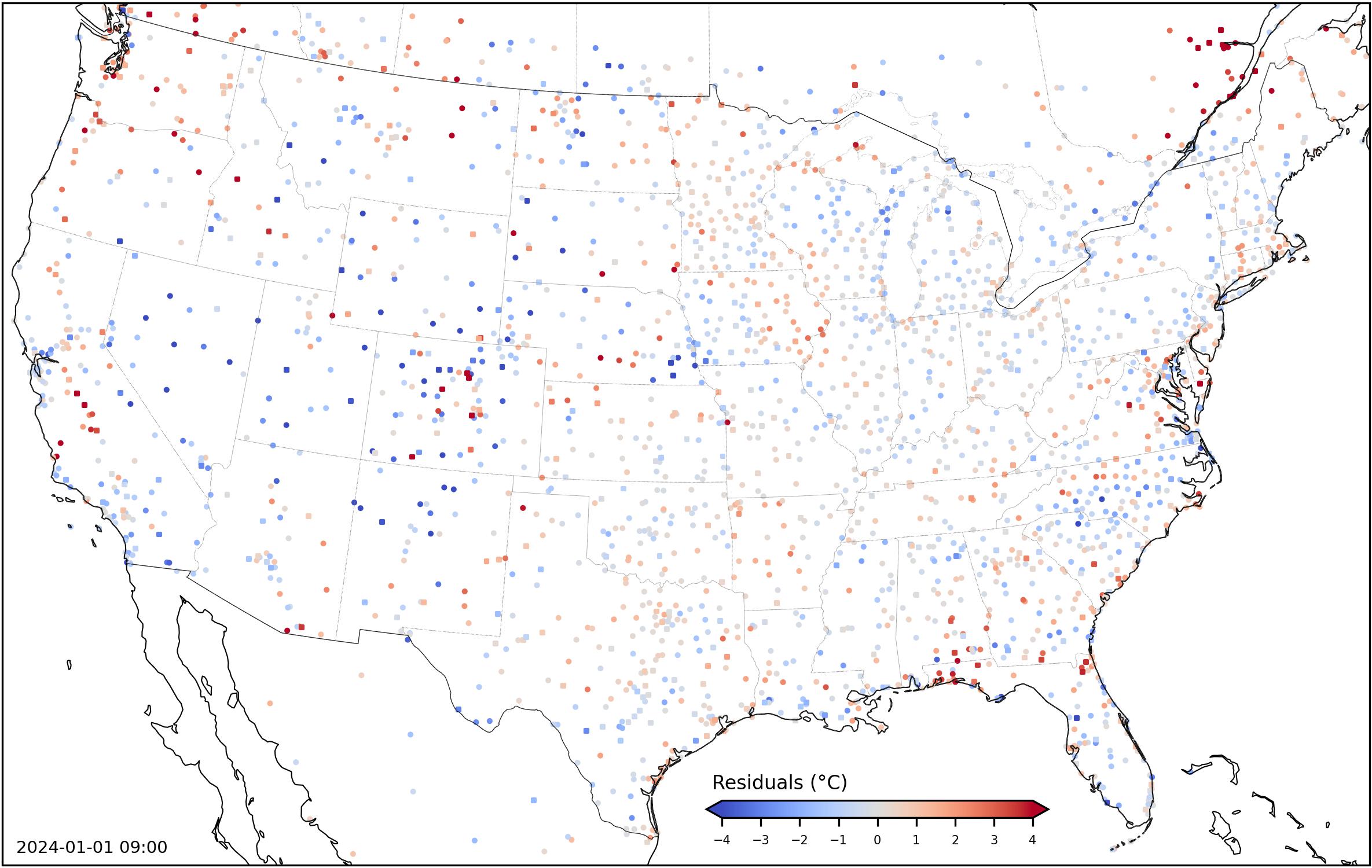} 
    \includegraphics[width=0.44\linewidth]{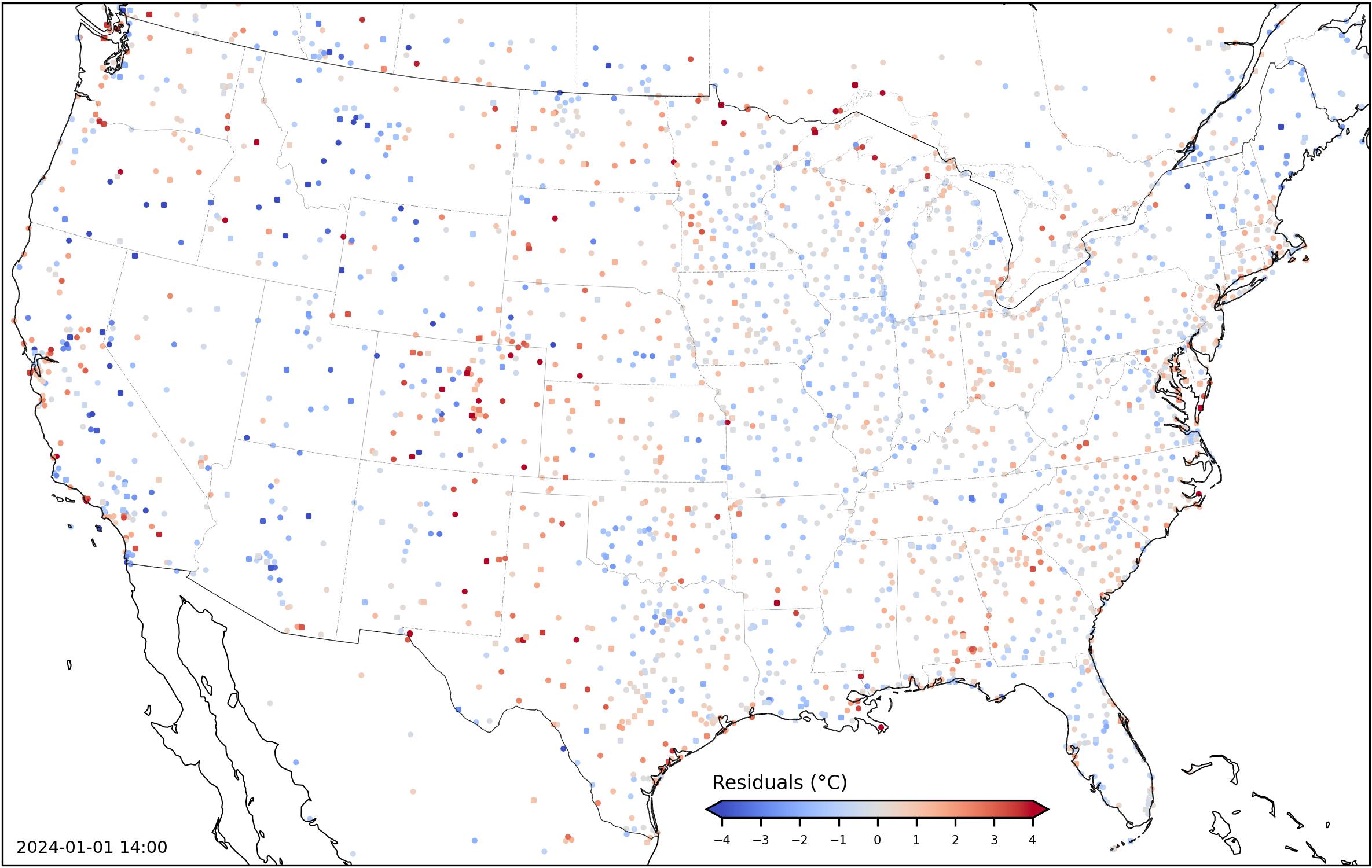}
    \caption{The cloud-obscured GOES-16 surface temperature data, the generated air temperature data, the in situ air temperature from meteorological stations, and the residuals at each station for 9:00 and 14:00 on January 1, 2024. Stations plotted as circles were used for training, and stations plotted as squares were used for testing.}
    \label{fig:maps2024b}
\end{figure*}

Similarly, we show two other instances on January 1, 2024, for 9:00 and 14:00 (UTC-5), in Figure~\ref{fig:maps2024b}. By comparing the estimates in July and January, we observe evident seasonal differences, such as the cooler temperature strips along the Rocky Mountains, the Peninsular Ranges in Southern California, and the Appalachian Mountains. By comparing the two air temperature prediction maps for summer and winter, we observe consistent, strong spatiotemporal patterns of air temperature data generated using the proposed method.

\begin{figure}[htbp]
    \centering
    \includegraphics[width=1.0\linewidth]{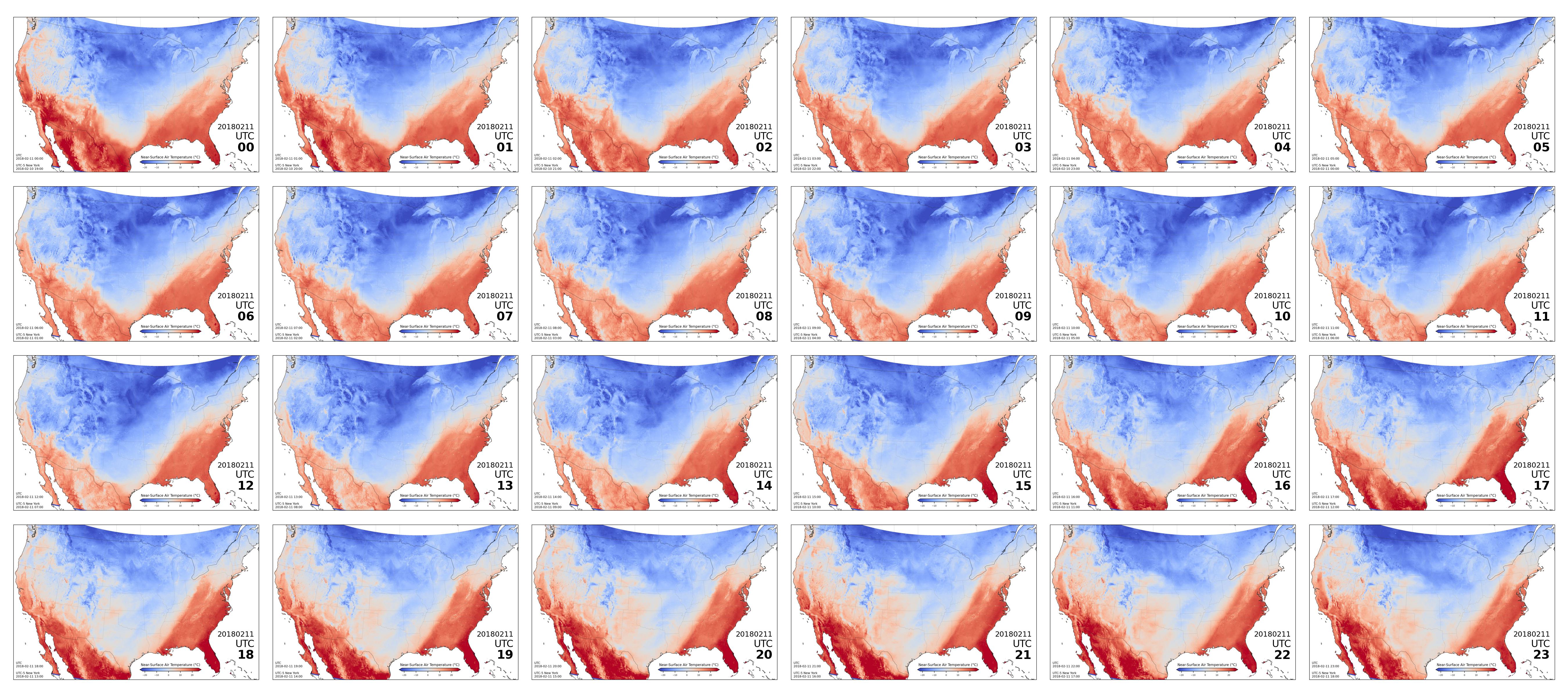}
    \caption{Prediction results on February 11, 2018 for all 24 hours.}
    \label{fig:pre20180211}
\end{figure}

\begin{figure}[htbp]
    \centering
    \includegraphics[width=1.0\linewidth]{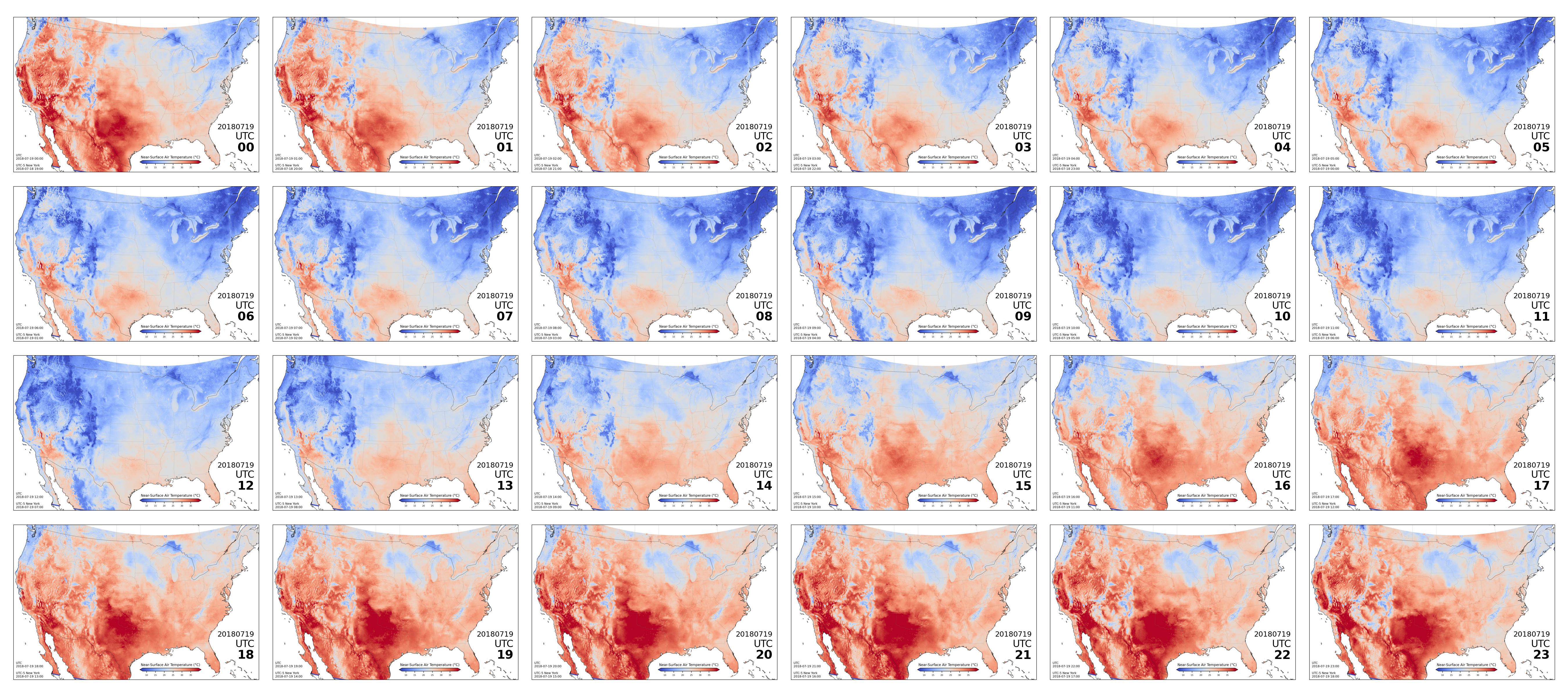}
    \caption{Prediction results on July 19, 2018 for all 24 hours.}
    \label{fig:pre20180719}
\end{figure}

We show two additional prediction maps, for all 24 hours, on February 11 and July 19, 2018 in Figures~\ref{fig:pre20180211} and ~\ref{fig:pre20180719} to show that the proposed approach successfully captured the changing temperature within the day. For best visualizations, we refer readers to the online GIF animations at \url{https://skrisliu.com/HourlyAirTemp2kmUSA/at2018042b.gif} and \url{https://skrisliu.com/HourlyAirTemp2kmUSA/at2018200b.gif}. 

\subsection{Uncertainty quantification}
\begin{figure}[htbp]
    \centering
    \includegraphics[width=0.8\linewidth]{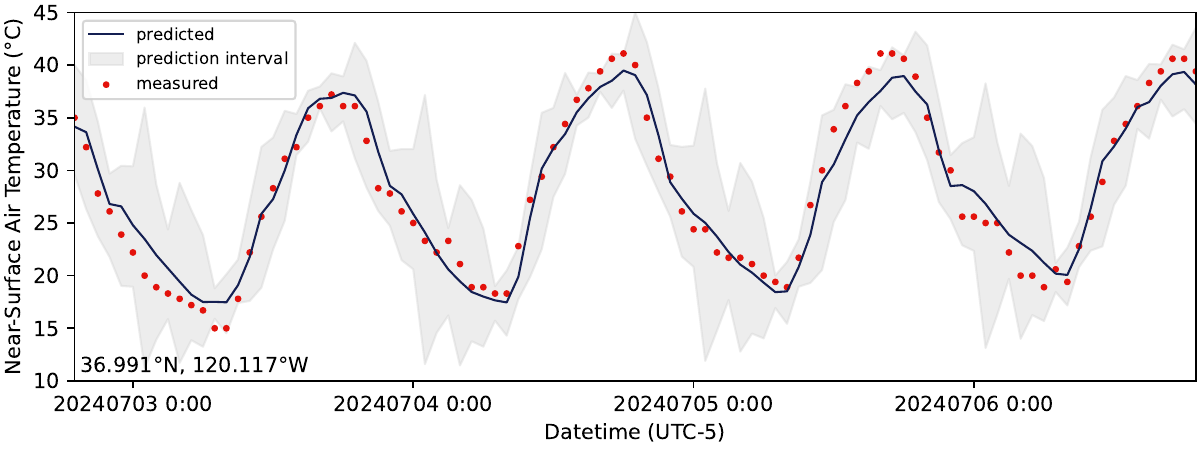}
    \caption{Predicted mean and intervals along with in situ air temperature data from one of the test stations.}
    \label{fig:Uncertainty}
\end{figure}

We show an example of the predicted mean and intervals from one of the test stations in Figure~\ref{fig:Uncertainty}. The predicted mean is plotted as a solid line, with the shaded area representing the prediction intervals. The in situ data—unseen during training—are plotted as red dots. Overall, the predictions successfully capture the diurnal temperature variations and are able to encompass the in situ values within the prediction intervals. Between 0:00 and 4:00 at midnight, the prediction intervals are slightly wider than at other times.

\subsection{Ablation analysis}
\subsubsection{Contribution of the Amplifier: surface temperature reconstruction}
\begin{figure}[htbp]
    \centering
    \includegraphics[width=0.5\linewidth]{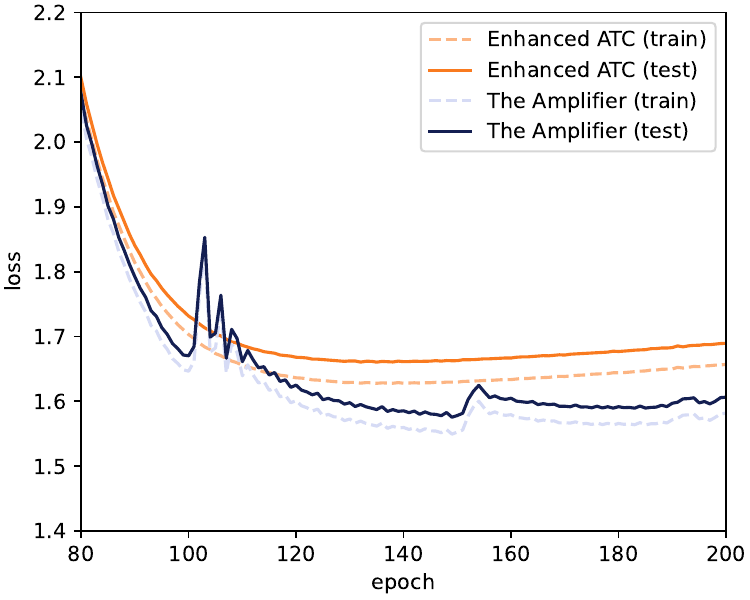}
    \caption{Comparison of L1 loss during training and testing between the proposed method (The Amplifier) and the enhanced ATC-only approach for surface temperature reconstruction.}
    \label{fig:loss}
\end{figure}

We first show the loss function comparison to highlight the effectiveness of the Amplifier model for surface temperature reconstruction (Figure~\ref{fig:loss}). Specifically, we compare the training and testing L1 losses between the baseline model (Enhanced ATC) and the proposed approach (the Amplifier). The example is conducted using the 2024 data, with 80\% of the data for training and 20\% for testing. This setup is designed to evaluate the impact of the spatiotemporal convolutional layers, which reduced the testing loss by an additional 0.1\degree C. The results underscore the critical role of incorporating coarse-resolution climate reanalysis data. By amplifying climate reanalysis data from coarse to fine resolution, the model better adheres to physical principles; and by using observational data to fit the Amplifier model, it consistently produces highly accurate estimates aligned with observations. Furthermore, the spatiotemporal convolutional layers contribute to additional performance gains by effectively capturing the inherent spatial and temporal dependencies in the surface temperature data.

\subsubsection{Performance comparison: air temperature transformation}
We conducted a series of ablation analyses to test the importance of input features and the effectiveness of the Air-Transformer of the proposed approach (Table~\ref{tab:performanceComparison}). The analysis is conducted over a subset area (east coast) in 2024. As shown in Table~\ref{tab:performanceComparison}, the proposed method (Amplifier Air-Transformer) has the best performance (MAE=1.47\degree C, RMSE=2.05\degree C), and the Air-Transformer improves the performance by 31\% (from 2.13\degree C to 1.47\degree C) in terms of MAE, compared to using multiple linear regression (MLR) with reconstructed surface temperature from the Amplifier. In terms of input features, ablation analysis shows that the addition of climate reanalysis data is the most critical, followed by spatial indexes (latitude and longitude), temporal index, and elevation-related variables (elevation and slope). For hourly air temperature estimation, the results show that having hourly information is more critical than having spatially varying features.

\begin{table}[htbp]
  \centering
  \caption{Performance comparison}
    \begin{tabular}{rrrr}
    \toprule
    \textbf{Method} & \textbf{MAE} (\degree C) & \textbf{RMSE} (\degree C) & \textbf{R$^2$} \\
    \midrule 
    Proposed w/o climate reanalysis   &   1.99  & 2.73 &  0.937 \\
    Proposed w/o hour index &    1.53   &   2.11    &   0.962     \\
    Proposed w/o elevation &   1.47     &    2.07   &  0.963      \\
    Proposed w/o lat/lon &  1.68     &  2.33     &   0.954     \\
    MLR w/ Amplifier      &   2.13    &  2.83     &    0.932    \\
    Proposed (Amplifier Air-Transformer) &   \textbf{1.47}    &  \textbf{2.05}     &    \textbf{0.964}     \\
    \bottomrule
    \end{tabular}%
  \label{tab:performanceComparison}%
\end{table}%

\section{Conclusion}
\label{sec:Conclusion}
High spatiotemporal resolution temperature data are essential for a wide range of environmental and societal applications. While satellites directly observe brightness temperature, which can be converted into LST using additional physical parameters, estimating near-surface air temperature--—a variable more directly linked to human and ecological systems—--remains challenging. This is largely due to the non-trivial relationship between LST and near-surface air temperature. 

In this study, we have proposed a data-driven, physics-guided deep learning approach (Amplifier Air-Transformer) to for hourly near-surface air temperature estimation at 2~km resolution. The proposed approach, consisting of two neural networks, streamline the estimation of all-weather near-surface air temperature from incomplete LST observations, through two neural networks. The Amplifier model, a convolutional network encoded with the annual temperature cycle and a linear term to downscale ERA5 surface temperature, enables LST reconstruction that is bounded by physical meanings from the Earth system models. The second neural network (Air-Transformer) fully explore the billions of surface temperature pixels and millions of air temperature records, leveraging the relationship among surface temperature, air temperature, and surface and near-surface properties, ultimately producing hourly estimation with high accuracy (RMSE=1.93\degree C). The approach is further enhanced with deep ensemble learning to provide non-parametric uncertainty quantification through a calibration process to generate reliable prediction intervals. 

Future work can strategically integrate additional physical principles to improve robustness under extreme conditions and enable simulation-based estimation. Furthermore, fusing observations from multiple--—potentially all--—satellite platforms holds great promise for advancing high-resolution temperature monitoring. These advancements will significantly enhance our ability to support a wide range of complex, real-world societal applications.

\section*{Acknowledgments}
S.K.L. was supported by a USC Dana and David Dornsife College of Letters, Arts and Sciences / Graduate School Fellowship at the University of Southern California. L.Z. was supported by the National Institutes of Health (NIH) under grant numbers P20HL176204, P30ES007048 and R01ES031590. The views and conclusions expressed in this study are those of the authors and do not necessarily represent the official policies or endorsements of the NIH. The authors sincerely appreciate the open data policies and practices of the National Oceanic and Atmospheric Administration (NOAA), the National Aeronautics and Space Administration (NASA), and the European Centre for Medium-Range Weather Forecasts (ECMWF), without which studies like this one would not have been possible.

\bibliographystyle{unsrt}
\bibliography{refs}

\end{document}